# Machine Learning-based Correlation of Charpy Impact Properties Between Sub-sized and Standard-sized Specimens for Nuclear Structural Materials

***Yugandhar Kasala Sreenivasulu*[1], *Isshu Lee*[2], *John W. Merickel*[2], *Fei Xu*[3,2], *Yalei Tang*[2], *Joshua E. Rittenhouse*[2], *Aleksandar Vakanski*[1,#], *Rongjie Song*[2,#]**

[1] University of Idaho, Department of Computer Science, Moscow, ID 83843, USA

[2] Idaho National Laboratory, Idaho Falls, ID 83415, USA

[3] Department of Computer Science, University of Texas at El Paso, El Paso, TX, United States

[#]Corresponding Authors: Aleksandar Vakanski: vakanski@uidaho.edu, Rongjie Song: Rongjie.Song@inl.gov

**Abstract**

Reliable correlations of Charpy impact test results between sub-sized and full-sized specimens are essential for structural integrity assessments, particularly in nuclear applications, where spatial constraints and limited material volume restrict specimen size. Although standards such as ASTM A370 and BS 7910 provide guidance on conversion methodologies, and numerous analytical correlation methods have been proposed in prior studies, these approaches generally have limited accuracy and their applicability is often constrained to specific materials, treatment conditions, and specimen geometries. In this study, a Machine Learning (ML)-based framework is proposed for correlating Charpy impact properties across specimen sizes. The proposed approach maps absorbed energy values across the full ductile-to-brittle transition region by applying a temperature shift combined with scaled residual projection, to align sub-sized test data with full-sized response. From the resulting temperature-energy profiles, the correlated values for upper shelf energy (USE) and ductile-to-brittle transition temperature (DBTT) are extracted by fitting data with a hyperbolic tangent model. The framework is validated using a dataset comprising 389 matched sub-sized and full-sized Charpy impact tests on SA533B steel. This ML-based approach demonstrates an improved correlation performance relative to conventional analytical methods, achieving $R^2$ values of 0.942 for USE and 0.892 for DBTT. The trained ML models do not require access to full-sized Charpy data during inference, making this approach suitable for material surveillance programs, accelerated irradiation testing, and other applications involving small-size Charpy impact testing.

## 1. Introduction

Charpy V-notch impact testing is a fundamental methodology for evaluating the fracture resistance and temperature-dependent impact toughness behavior of ferritic steels used in reactor pressure vessels (RPV) [1−3]. The primary parameters obtained from Charpy tests performed within a range of temperatures include the upper shelf energy (USE) and the ductile-to-brittle transition temperature (DBTT). The USE is defined as the plateau value of absorbed impact energy measured at sufficiently high temperatures where fracture occurs predominantly by ductile mechanisms. The DBTT is defined as the temperature corresponding to the inflection point of the sigmoidal absorbed energy-temperature curve, representing the transition from brittle to ductile fracture behavior. These parameters serve as important indicators for quantifying a material's capacity to withstand dynamic loading under unirradiated and neutron-irradiated service conditions. Within the material surveillance programs in nuclear power plants, long-term monitoring of USE and DBTT through periodic testing of archived samples fabricated from representative materials is indispensable for ensuring the structural integrity of reactor components over their licensed operational lifespan and extended operation [4,5]. However, challenges with the availability of archived irradiated material limit the extent and flexibility of these programs, particularly in the context of plant life extension. A viable approach to expanding material surveillance programs has involved the use of sub-sized specimens, which enables multiple smaller samples with reduced dimensions to be extracted from previously tested full-sized specimens [6−8]. In addition, fractured or previously tested Charpy specimens can be reconstituted to produce new test specimens for additional impact testing [9]. An alternative approach for characterizing irradiation-induced embrittlement involves accelerated irradiation campaigns conducted in test reactors, where materials are subjected to significantly higher neutron fluxes than under typical service conditions in surveillance programs [7]. In this context, the adoption of sub-sized specimens has also emerged as a practical and efficient means of expanding data acquisition, due to the spatial constraints and limited irradiation volume available within test reactor environments.

In addition, the utilization of small-size specimen testing techniques has gained increasing importance in other settings, where standard-sized specimens are impractical or infeasible due to constraints related to limited material availability, complex component geometries, or restrictive experimental environments. Notable applications include residual life assessment of in-service components through minimally invasive sampling [10], high-throughput alloy development using miniature specimens for rapid prototyping and screening [11,12], and localized mechanical characterization in spatially constrained regions, such as welded joints or heat-affected zones [13].

Despite the apparent benefits and advantages of conducting tests with sub-sized Charpy specimens, their reduced dimensions introduce deviations in fracture behavior, due to altered underlying fracture mechanisms in comparison to standard full-sized specimens. These differences stem from geometry-dependent factors, such as reduced ligament size, modified constraint conditions at the notch root, and changes in stress triaxiality, which collectively influence energy absorption and ductile-to-brittle transition behavior [2,5−8,14]. Broadly referred to as *size effects*, these phenomena can lead to systematic underestimation or overestimation of Charpy impact properties such as DBTT and USE with respect to full-sized specimens. Consequently, mechanical responses from sub-sized specimens cannot be directly interpreted as equivalent to those of full-sized counterparts, and necessitate the development and application of reliable correlation methodologies [15].

Extensive research effort has been dedicated over the past several decades to developing analytical approaches for correlating Charpy impact properties obtained from sub-sized specimens with those from standard full-sized specimens. These efforts have yielded a multitude of empirical and semi-empirical methods, including approaches for direct estimation of USE and DBTT from sub-sized test data, as well as indirect approaches that first normalize absorbed energy values across temperature ranges prior to extracting the correlated parameters. For USE correlation in particular, several scaling laws have been proposed based on specimen geometry and dimension-dependent factors, including specimen thickness, ligament size [16–18], span length [19], and stress concentration factors based on the notch effect [5,20,21]. Additional correlation methods include linear regression models utilizing power-law scaling relationships [6,8,15,22], fracture mechanics-based formulations [5,21], and finite element simulations that account for differences in constraint and energy dissipation mechanisms [14,23]. Similarly, DBTT correlation approaches have been developed using constant temperature shift assumptions between sub-sized and full-sized specimens [6,24,25], temperature adjustments based on stress concentration factors [19,20], and data-driven models employing regression techniques [8].

Current correlation methods exhibit limitations in capturing the complex relationships between sub-sized and full-sized Charpy impact properties across a broad range of materials and testing conditions. First, analytical models lack mathematical flexibility to incorporate critical factors that influence Charpy behavior, such as material treatment history, irradiation effects, microstructural variability, specimen orientation relative to the production process, and so on. For example, in this study, the USE for full-sized specimens of the same material with identical nominal dimensions and geometry varies significantly, ranging from 51.8 J to 278 J. Correlation methods based solely on specimen geometry cannot adequately reflect these important effects. Second, most existing models are based on linear or power-law formulations, which represent approximations of the underlying nonlinear and potentially multivariate dependencies between Charpy parameters. Third, empirical and semi-empirical methods, such as methods based on stress concentration factors or experimentally calibrated parameters, were typically developed for specific materials processed under controlled conditions. As a result, these methods often necessitate re-calibration of correlation parameters when applied to new materials, and even to different heat treatment or processing conditions of the same alloy, thereby limiting their generalizability. Fourth, the development of traditional correlation methods was often constrained by limited experimental datasets, contributing to high variability and uncertainty in the estimated properties. Moreover, analytical models that attempt to correlate absorbed energies across the entire ductile-to-brittle temperature range often fail to accurately capture the influence of reduced specimen dimensions, leading to overestimation of absorbed energies in the lower shelf region and underestimation in the upper shelf region [22]. These limitations emphasize the need for advanced, flexible, and generalizable computational frameworks to reliably correlate Charpy properties between sub-sized and full-sized specimens.

This study presents a data-driven framework employing Machine Learning (ML) models to correlate full-sized Charpy impact properties and sub-sized specimen data. The proposed methodology involves predicting absorbed energies across the full temperature range using ML modeling, followed by indirect estimation of the USE and DBTT parameters by fitting a hyperbolic tangent (tanh) function to the predicted values. To compile matched sub-sized and full-sized datasets for this task, we collected data from published sources that report both sub-sized and full-sized Charpy impact test results under identical specimen processing and testing conditions. For transforming the sub-sized Charpy data into full-sized equivalents, we applied a scaling approach that preserves the variability of temperatures and absorbed energies around the sub-sized curve and aligns the curve shape in the transition region with that of the corresponding full-sized data. The resulting scaled temperature-energy values served as targets for training a supervised ML model, with the original sub-sized temperature-energy data serving as input features. Experimental validation was conducted using a dataset of 389 Charpy impact test records for SA533B steel [26]. The results demonstrate improved performance compared to existing analytical correlation methods, with a coefficient of determination $R^2$ of 0.942 and Pearson correlation coefficient $r$ of 0.974 for USE, and 0.892 and 0.948 for DBTT, respectively. The improvements are attributed to the ML model's ability to capture complex dependencies across 37 input variables that encompass chemical composition, thermal treatment, cooling method, irradiation exposure, microstructural characteristics, test conditions, specimen geometry, and other relevant parameters.

To the best of our knowledge, this is the first application of ML aimed at addressing the size effects correlation in Charpy impact testing. While numerous previous studies have successfully applied ML techniques to predict Charpy impact properties, these efforts have predominantly focused on standard full-sized specimens [27−30]. Likewise, other research has explored ML-based property predictions for sub-sized specimens in different mechanical tests. Examples include models developed for estimating the fatigue life of additively manufactured sub-sized specimens [31,32], and ML-based predictions of yield strength, ultimate tensile strength, uniform elongation, and total elongation from sub-sized tensile test data [33]. In another related work, Pan et al. [34] employed ML techniques to correlate force-displacement curves from small punch tests with stress-strain curves obtained from sub-sized tensile specimens for pressure vessel steels. These efforts have focused on specimen size effects in fatigue, tensile, or small punch testing. Instead, our work addresses the size correlation challenge in Charpy impact testing, where energy absorption and ductile-to-brittle transition behavior are significantly influenced by specimen geometry and size.

The main outcomes of this study are as follows:

- Developing a novel ML-based correlation framework that relates sub-sized Charpy specimen properties to full-sized specimen behavior using temperature shifting and scaled residual projection.
- Providing a comprehensive comparative evaluation that demonstrates increased agreement of the proposed ML framework with Charpy impact test results for both USE and DBT with respect to established analytical correlation models.
- Introducing uncertainty quantification strategies that estimate confidence and prediction intervals for USE and DBTT, addressing the reliability of the correlated properties.

## 2. Background

### 2.1. Charpy Impact Test

The Charpy impact test is a standardized method for evaluating the impact toughness of materials under dynamic loading conditions. The Charpy impact test setup is depicted in Figure 1(a), where a notched specimen is placed horizontally on supports, and a pendulum hammer, released from a fixed height, strikes the specimen opposite the notch and fractures it. The energy absorbed by the material during fracture is determined by measuring the height to which the pendulum swings after breaking the specimen. The absorbed energy reflects the material's ability to resist impact and deformation.

The experimental procedure standardized in ASTM E23 [1] involves conducting Charpy impact tests at various temperatures. The measured values of impact energy plotted versus temperature are used to establish a pattern of energy absorption. A typical ductile-to-brittle transition pattern is shown in Figure 1(b) and consists of three regions: an upper shelf region with high, nearly constant impact energy, a transition region characterized by a steep variation in energy absorption, and a lower shelf region with low, nearly constant impact energy. USE and DBTT, as key material parameters extracted from impact test data, are essential for assessing material performance under varying conditions and identifying the material's suitability for specific operating conditions.

### 2.2. Specimen Size Effect in Charpy V-Notch Impact Test

It is important to clarify that in this study we use the term "sub-sized Charpy specimen" to refer broadly to any Charpy specimen with reduced dimensions relative to the standard full-sized geometry with 10 mm × 10 mm × 55 mm dimensions (shown in Figure 1(c)). Some standards [1,35] use the term "sub-sized" specifically for Charpy specimens with reduced thickness and all other dimensions preserved, whereas Charpy specimens with additional dimensional reductions of width, length, or notch dimensions are referred to as "miniature" specimens [22,25].

Size effects in Charpy V-notch (CVN) impact testing introduce significant variations in measured absorbed energies across different temperatures. Sub-sized specimens, due to their reduced dimensions, have different fracture behavior and energy absorption capacity compared to full-sized specimens, even when tested under identical material and environment conditions. These differences result in lower absorbed energy values, and also cause a shift in the Charpy impact properties, where USE is reduced and DBTT is typically shifted to lower temperatures with decreasing specimen thickness.

Louden [19] attributes the size effects in Charpy impact testing primarily to changes in the stress state at the notch tip that occur as specimen dimensions are reduced. In full-sized specimens, the constraint conditions near the notch root are predominantly characterized by plane strain, which increases triaxial stress and promotes brittle fracture, leading to higher DBTT. When specimen size is reduced, the proportion of the cross-section subject to plane stress increases, which facilitates out-of-plane deformation, allowing the material to undergo ductile fracture. As a result, sub-sized specimens appear more ductile at a given temperature, yielding lower DBTT values. Louden also notes that USE decreases with specimen size due to the smaller fracture volume involved in energy absorption during impact testing.

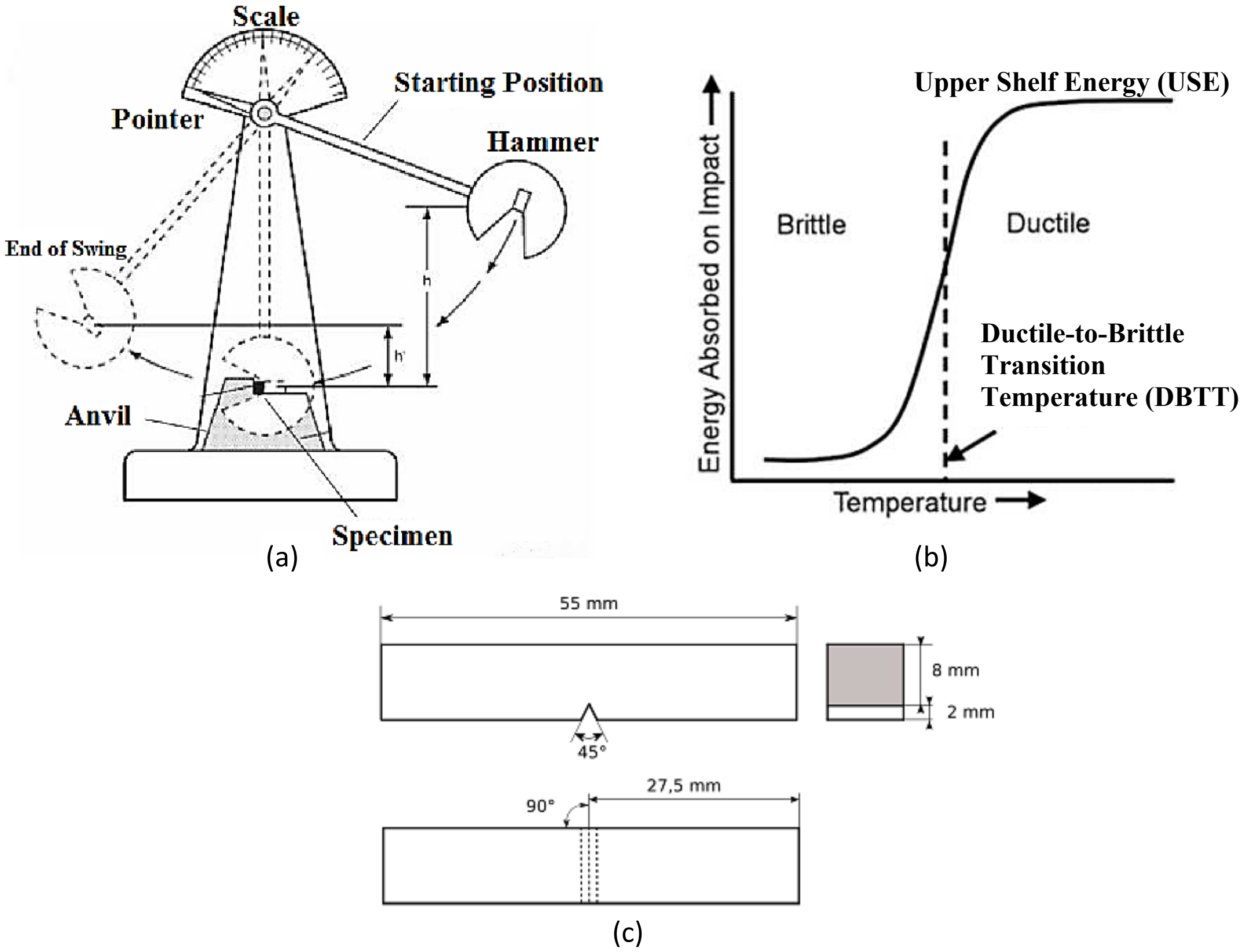

**Figure 1.** (a) Illustration of Charpy impact test setup [35]. (b) Ductile-to-brittle transition curve. (c) Standard Charpy V-notch impact test specimen.

Wallin [15] expanded earlier postulations on size effects in Charpy impact testing by introducing a perspective based on the evolution of competing fracture micromechanics as specimen geometry decreases. Specifically, the fracture surface of a CVN specimen typically comprises a central flat fracture region and external shear lips, corresponding to ductile fracture regions along the sides of the specimen. The shear lips region increases with crack propagation until it reaches a saturation level dependent on the material's tearing resistance. As specimen dimensions decrease, the proportion of the fracture surface occupied by the shear lips increases, and in very thin specimens, shear fracture may dominate the entire surface. Because the energy required to form shear fracture is different than that for flat fracture, this shift in fracture morphology alters the measured absorbed energy, typically resulting in higher energy measurements in the lower shelf region, and lower energy measurements in the upper shelf region. Moreover, the transition region between the lower and upper shelf presents a combination of ductile and brittle fracture micromechanics resulting in a complex, non-trivial size effect. Wallin's work highlights the significance of fracture surface mechanisms in addition to stress state effects, offering a more comprehensive framework for understanding size effects in Charpy testing. On the other hand, the findings also emphasize the inherent challenges in interpreting sub-sized Charpy data, as specimen geometry introduces region-specific effects across the transition curve, necessitating robust conversion methodologies to reliably estimate equivalent full-size Charpy properties.

### 2.3. Analytical Models for Correlating Charpy Impact Properties

This section provides a brief review of some analytical methods for correlating USE and DBTT between sub-sized and full-sized Charpy specimens.

*2.3.1. Upper Shelf Energy Correlations*

Most analytical methods for correlating the USE involve the use of normalization factors which adjust the measured USE values between full-sized and sub-sized Charpy specimens to account for difference in specimen geometry, following the relation:

$$\frac{\mathrm{USE}_f}{\mathrm{NF}_f} = \frac{\mathrm{USE}_s}{\mathrm{NF}_s} \tag{1}$$

where $\mathrm{USE}_f$ and $\mathrm{USE}_s$ are the USE values of full-sized and sub-sized specimens, and $\mathrm{NF}_f$ and $\mathrm{NF}_s$ denote the corresponding normalization factors, respectively. From this relationship, the full-sized USE obtained from sub-sized specimen, denoted $\mathrm{USE}_f$, can be calculated as:

$$\mathrm{USE}_f = \mathrm{USE}_s \frac{\mathrm{NF}_f}{\mathrm{NF}_s} \tag{2}$$

Several normalization factors have been proposed in the literature, either derived analytically based on geometric dependencies of USE, or empirically from Charpy impact test data.

Lucas et al. [16,17] proposed a correlation approach based on observations that absorbed energy in the upper shelf region increases proportionally with the fracture volume. To account for differences in specimen geometry, the authors introduced a normalization factor proportional to $Bb^2$, where $B$ denotes the specimen thickness, $b = W - A$ is the ligament size, $W$ is the specimen width, and $A$ is the notch depth. Corwin et al. [18] proposed a similar normalization approach using a fracture volume factor given by $(Bb)^{3/2}$. Their analysis demonstrated that this scaling improves the correlation in higher energy ranges dominated by ductile fracture. A shortcoming of these methods is that volume-based normalization does not fully account for differences in constraint conditions and notch effects, which can reduce the accuracy of the correlation.

Louden et al. [19] introduced an approach that stems from observations that reduced specimen sizes alter the constraint at the notch root, making sub-sized specimens appear more ductile, leading to inconsistencies in energy correlation. To address this, they suggested a normalization factor $Bb^2/LK_t$, where $L$ is the span length (horizontal distance between the two supports on the Charpy testing machine), and $K_t$ represents the stress concentration factor that is a function of the ligament size and the radius of the notch. By accounting for both fracture volume and notch effects, the method aims to improve the assessment of USE across specimen sizes, particularly for low-ductility materials, where constraint effects are more pronounced. However, its accuracy is reduced for high-toughness materials with USE above 150 J, where deformation mechanisms, such as work hardening, become dominant and are not fully captured by the normalization factor.

Kayano et al. [20] developed a normalization method specifically aimed at correlating USE for miniaturized Charpy specimens in the millimeter-scale thickness range. The scaling approach $Bb^2/K_t'$ employs a modified stress concentration factor $K_t'$ that incorporates both elastic stress concentration and plastic constraint effects. It is calculated as $K_t' = K_t \times Q$, where $Q$ is the plastic stress concentration factor, given by $Q = 1 + \pi/2 - \theta/2$ based on notch angle $\theta$. The method achieved an improved consistency of energy-temperature curves for specimens as thin as 1.5 mm. However, for miniature specimens with 1.0 mm thickness, the pronounced notch-root plasticity and reduced ligament size led to deviations in the normalized USE values. Building on these ideas, Schubert et al. [5] proposed a related approach using a normalization factor $Bb^2/LK_t''$. The modified stress concentration factor $K_t''$ in Shubert et al. [5] is also calculated as $K_t \times Q$, however, it uses adjusted values of the stress concentration factor $K_t$. This formulation was validated for materials with USE values below 100 J, and showed improved consistency of correlated sub-sized USE values.

Kumar et al. [21] proposed a correlation method that decomposes the total absorbed energy into two components: crack initiation energy ($\Delta\mathrm{USE}$) and crack propagation energy ($\mathrm{USE}_p$). They argued that previous normalization methods fail to capture the differences in fracture mechanics between the initiation and propagation phases, particularly in sub-sized specimens, where constraint conditions vary significantly. To address this, the authors employed both notched and pre-cracked Charpy specimens, directly measuring $\mathrm{USE}_p$ from the pre-cracked samples and determining $\Delta\mathrm{USE}$ as the difference between the total energy and $\mathrm{USE}_p$. This approach improved the consistency of energy correlations between sub-sized and full-sized specimens, especially for steels with low to moderate ductility. Limitations included relying on testing pre-cracked specimens, which increases experimental complexity, and the reduced applicability to highly ductile materials, where initiation energy dominates, and the partitioning into $\Delta\mathrm{USE}$ and $\mathrm{USE}_p$ becomes less effective.

Sokolov et al. [6] introduced a fracture mode-based normalization method that incorporates the percentage of shear fracture on the fracture surface as a direct indicator of fracture mode. The proposed energy correction is expressed as $\mathrm{NF}_{\mathrm{brittle}}(1-S)+\mathrm{NF}_{\mathrm{ductile}}\,S$, where $\mathrm{NF}_{\mathrm{brittle}}$ and $\mathrm{NF}_{\mathrm{ductile}}$ are empirical normalization factors corresponding to brittle and ductile fracture contributions, and $S$ is the percentage of shear fracture. This formulation improves the correlation of absorbed energy values across specimen sizes, by accounting for the actual fracture behavior rather than relying solely on specimen geometry. However, this method depends on the accurate measurement of the shear area, and the need to calibrate $\mathrm{NF}_{\mathrm{brittle}}$ and $\mathrm{NF}_{\mathrm{ductile}}$ for each material can limit its general applicability.

Uehira et al. [8] proposed a correlation method based on a power law normalization term, $m(Bb)^n$, where $m$ is a material-dependent constant, and $n$ is an empirically determined exponent related to the material's ductility. The exponent $n$ is estimated by fitting a linear regression model to USE values of both sub-sized and full-sized specimens, while the typical values for $m$ range between 1.2 and 1.7. This method offers a relatively straightforward way of estimating full-sized USE from sub-sized data. A limitation is the requirement for experimental data from specific materials and treatment conditions to calibrate the coefficients $m$ and $n$.

Wallin et al. [37] presented an updated empirical method that employs a normalization approach based on an exponential relationship of the form $(Bb)/b^m$, where $m$ is an empirically derived exponent. This formulation captures the nonlinear reduction in USE with decreasing ligament size, and avoids the need for detailed fracture surface measurements. As an empirical method, however, it requires calibration against reference Charpy datasets and verification for new materials.

The described USE correlation methods are listed in Table 1.

**Table 1.** Analytical methods for correlating USE between sub-sized and full-sized Charpy specimens.

| Authors | Estimated Full-size USE | Approach |
|---|---|---|
| Lucas et al. [16,17] | $\mathrm{USE}_s \cdot \left(B_f b_f^2\right)/\left(B_s b_s^2\right)$ | Normalization |
| Corwin et al. [18] | $\mathrm{USE}_s \cdot \left(B_f b_f\right)^{3/2}/\left(B_s b_s\right)^{3/2}$ | Normalization |
| Louden et al. [19] | $\mathrm{USE}_s \cdot \left(B_f b_f^2/L_f K_{t,f}\right)/\left(B_s b_s^2/L_s K_{t,s}\right)$ | Normalization with stress concentration factor |
| Kayano et al. [20] | $\mathrm{USE}_s \cdot \left(B_f b_f^2/K'_{t,f}\right)/\left(B_s b_s^2/K'_{t,s}\right)$ | Normalization with modified stress concentration factor |
| Schubert et al. [5] | $\mathrm{USE}_s \cdot \left(B_f b_f^2/L_f K''_{t,f}\right)/\left(B_s b_s^2/L_s K''_{t,s}\right)$ | Normalization with modified stress concentration factor |

| Kumar et al. [21] | $\Delta\text{USE} + \text{USE}_p$ | Sum of crack initiation and crack propagation energy |
|---|---|---|
| Sokolov et al. [6] | $\text{USE}_s \cdot (\text{NF}_{\text{brittle}}(1 - S) + \text{NF}_{\text{ductile}}S)$ | Empirical sum of ductile fracture and brittle fracture |
| Uehira et al. [8] | $\text{USE}_s \cdot \left(m\left(B_f b_f\right)^n\right)/(m(B_s b_s)^n)$ | Empirical power-law correlation |
| Wallin et al. [37] | $\text{USE}_s \cdot \left(b_f/b_s\right)^m \cdot \left(B_f b_f\right)/(B_s b_s)$ | Empirical power-law correlation |

*2.2.2 Ductile-to-Brittle Transition Temperature Correlations*

Most of the analytical methods for correlating DBTT values utilize an empirically determined temperature shift having the form:

$$\text{DBTT}_{f,cor} = \text{DBTT}_s + \text{C} \quad (3)$$

where C is a constant that accounts for differences in geometry or constraint effects between specimen sizes.

Towers [24] proposed a simple empirical method to estimate the DBTT shift based on the reduction in constraint with decreasing specimen thickness. The method uses a quadratic relationship for the temperature shift, $0.7(10 - B)^2$, where the squared term captures the non-linear decrease in DBTT for thinner specimens. This method relies solely on geometric scaling without considering material properties or fracture mechanisms.

Louden et al. [19] introduced a correlation method that accounts for constraint changes when evaluating embrittlement effects, including those induced by irradiation or heat treatment. The method employs a normalization factor by dividing the measured DBTT by a constraint parameter $\sigma' = K_t 3P_m L/2Bb^2$, where $P_m$ is the maximum force recorded during the instrumented Charpy test. Limitations of this approach include the requirement for the determination of $P_m$, and the assumption of a size-independent embrittlement shift, which may not be applicable to heterogeneous materials or miniature specimens.

Kayano et al. [20] developed an analytical method for estimating full-sized DBTT from miniaturized Charpy specimens with thicknesses of 1.5 mm and 1.0 mm. The approach utilizes a normalization factor based on the inverse square root of a modified stress concentration factor, $K_t'^{1/2}$. While the method achieved accurate correlation for 1.5 mm specimens, it tended to underestimate DBTT for 1.0 mm specimens due to enhanced plastic constraint effects, not fully captured by the normalization factor.

Kurishita et al. [38] identified the ratio between notch depth and notch root radius, $A/\rho$, as a dominant parameter influencing DBTT shifts. Their analysis showed that DBTT increases with increasing $A/\rho$, and that a linear relationship exists between DBTT and this ratio. They demonstrated that when $A/\rho$ exceeds approximately 15, DBTT values for sub-sized specimens can approximate those of full-sized specimens, with errors typically smaller than 10 K. This method has been primarily validated on unirradiated low-activation ferritic steels, and may require recalibration for broader applications.

Sokolov et al. [6] proposed a DBTT correction method based on a logarithmic temperature shift: $98 - 15.1 \cdot \ln(Bb^2)$. This formulation is based on statistical fracture mechanics, where cleavage fracture probability depends on the stressed volume and defect population, and accounts for the effects of reduced constraint and sampling volume in smaller specimens. The method showed good agreement with full-sized transition temperature data across various materials. Building on this approach, Lucon et al. [25] developed a similar empirical correlation:$92.7 - 14.34 \cdot \ln(Bb^2)$, which improved the accuracy for RPV steels across various sub-sized geometries, including half-sized, two-thirds-sized, and KLST specimens. Shortcomings of these methods include potential inaccuracies in ultra-small specimens and reduced applicability to materials with high ductility or significant plasticity contributions.

Uehira et al. [8] developed an empirical DBTT correlation method to account for both specimen size and notch effects in ferritic-martensitic stainless steels. The authors proposed the following form for the correlated DBTT: $115 \cdot \log_{10}(BK_t) - 158$, where $K_t$ is the elastic stress concentration factor. While the method showed good accuracy for PNC-FMS and similar steels, it is limited by its empirical nature.

Wallin et al. [37] proposed an alternative approach based on a DBTT shift given by: $51.4 \cdot \ln(2(B/10)^{0.25} - 1)$. While effective for conventional ferritic materials, the accuracy of this method may be reduced for miniature specimens or materials exhibiting high ductility.

The described approaches for DBTT correlation are listed in Table 2.

**Table 2.** Analytical methods for correlating DBTT between sub-sized and full-sized Charpy specimens.

| Authors | Estimated Full-size DBTT | Explanation |
|---|---|---|
| Towers [24] | $\mathrm{DBTT}_s + 0.7\,(10 - B_s)^2$ | Temperature shift by a constant |
| Louden et al. [19] | $\mathrm{DBTT}_s\big(2B_f b_f^2/K_{t,f}3P_{m,f}L_f\big)/(2B_s b_s^2)/K_{t,s}3P_{m,s}L_s$ | Normalization with stress concentration factor |
| Kayano et al. [20] | $\mathrm{DBTT}_s {K'_{t,f}}^{1/2}/{K'_{t,s}}^{1/2}$ | Normalization with modified stress concentration factor |
| Kurishita et al. [38] | $\text{if } (A/\rho) \sim 15 \ \rightarrow \mathrm{DBTT}_f$ | Use of the notch depth to root radius ratio |
| Sokolov et al. [6] | $\mathrm{DBTT}_s + 98 - 15.1 \cdot \ln(B_s b_s^2)$ | Temperature shift by a constant |
| Lucon et al. [25] | $\mathrm{DBTT}_s + 92.7 - 14.34 \cdot \ln(B_s b_s^2)$ | Temperature shift by a constant |
| Uehira et al. [8] | $115 \cdot \log_{10}\big(B_s K_{t,s}\big) - 158$ | Empirical correlation |
| Wallin et al. [37] | $\mathrm{DBTT}_s - 51.4 \cdot \ln\Big(2\big(B_s/B_f\big)^{0.25} - 1\Big)$ | Temperature shift by a constant |

## 3. Materials and Methods

### 3.1. Dataset of Charpy Impact Properties for Sub-sized Specimens

To support the development of analytical and ML models for correlating Charpy impact properties across specimen sizes, we compiled an open-access dataset for nuclear structural materials [26]. The dataset was generated through an extensive literature review of 109 peer-reviewed publications and technical reports. It encompasses a broad range of materials, including SA508, SA533B, 20MnMoNi55, A302B, SS316, and 15Cr2MoV-A. In total, the dataset contains 4,937 individual Charpy impact test records, corresponding to 293 unique Charpy curves. It is publicly available through the Materials Cloud repository [39], and a detailed description of the dataset is provided in an article published in Scientific Data [26]. Each record includes information grouped into categories related to material composition, thermal and irradiation conditions, manufacturing and treatment details, specimen dimensions, test conditions, and Charpy impact properties. All entries were manually validated by materials science experts, and importantly, the dataset covers both full-sized and sub-sized specimens, making it well-suited for a systematic evaluation of size effects.

In this study, we utilize a subset of the Charpy impact dataset for the SA533B alloy. To validate the investigated correlation models, we selected only the sub-sized specimen data for which corresponding experimental results from full-sized specimens are available in the original publications [5,40–42]. This filtering process produced 389 sub-sized Charpy test records derived from 27 distinct sub-sized specimen sets of SA533B. The corresponding full-sized specimens are used for evaluating the correlation methods. Important parameters for the specimens used in this study are shown in Table 3. In addition, more details for both the used sub-sized specimens and the full-sized counterparts are provided in Table A1 in Supplementary Material.

**Table 3.** Dimensions and geometry parameters of specimens used in this study, including the stress concentration factor $K_t$ from Louden et al. [19], modified stress concentration factor $K_t'$ from Kayano et al. [20][1], and modified stress concentration factor $K_t''$ from Schubert et al. [5].
[1]The values for $K_t'$ are from reference [5] which investigates SA533B.

| Size | Thickness $B$ (mm) | Width $W$ (mm) | Length $L_0$ (mm) | Span $L$ (mm) | Ligament Size $b$ (mm) | Notch Depth (mm) | Notch Angle | Notch-Root Radius, ρ (mm) | $K_t$ | $K_t'$ | $K_t''$ |
|---|---|---|---|---|---|---|---|---|---|---|---|
| **Full** | 10 | 10 | 55 | 40 | 8 | 2 | 45° | 0.25 | 4.8 | 4.81 | 10.49 |
| **Half** | 5 | 5 | 23.6 | 20 | 4.24 | 0.76 | 30° | 0.08 | 7.6 | 6.09 | 17.23 |
| **Third** | 3.33 | 3.33 | 23.6 | 20 | 2.82 | 0.51 | 30° | 0.08 | 6.2 | 5.03 | 14.23 |
| **Quarter** | 2.5 | 2.5 | 20 | 16 | 2.12 | 0.38 | 30° | 0.08 | 5.7 | 4.53 | 11.7 |
| **Mini** | 1.5 | 1.5 | 20 | 16 | 1.17 | 0.33 | 30° | 0.08 | 5.4 | 5.11 | 11.1 |

### 3.2. Machine Learning Approach for the Correlation of Charpy Impact Properties

#### *3.2.1. Data Transformation*

To train ML models for correlating sub-sized to full-sized Charpy impact properties, the data from each sub-sized specimen set was transformed to align with the corresponding full-sized specimen set. The Charpy impact test data for a sub-sized specimen are denoted $(T_s^1, E_s^1), (T_s^2, E_s^2), \ldots$ $(T_s^N, E_s^N)$, where $T_s^i$ represents the test temperature, $E_s^i$ is the measured absorbed energy for the $i^{\text{th}}$ test, and $N$ is the number of Charpy impact tests performed on that material, typically ranging from 8 to 12. Analogously, the data for the corresponding full-sized specimen set, produced under the same manufacturing and heat treatment conditions and tested using the same test protocol are denoted $\left(T_f^1, E_f^1\right), \left(T_f^2, E_f^2\right), \ldots, \left(T_f^M, E_f^M\right)$, where $M$ is the number of tests.

The absorbed energy data is commonly modeled using a sigmoidal function, with the hyperbolic tangent (tanh) function being one of the most widely used formulations [8,36,43]. A standard representation is:

$$f(T) = \frac{\text{USE}}{2}\left(1 + \tanh\frac{T - \text{DBTT}}{k}\right) \tag{4}$$

where $f(T)$ is the absorbed energy at temperature $T$, $\text{USE}$ and $\text{DBTT}$ are the Charpy impact properties, and $k$ is a parameter controlling the slope of the transition between the upper shelf and lower shelf. The tanh formulation has also been widely used to model other temperature-dependent Charpy transition properties, such as lateral expansion and shear fracture appearance, which exhibit a similar sigmoidal transition behavior. In addition to this formulation, alternative versions of the tanh model that incorporate the lower shelf energy have also been employed [27], as well as other sigmoidal models derived from the Burr and Weibull distributions [27].

The fitted curves for a representative sub-sized specimen configuration and the corresponding full-sized specimen configuration in the dataset, $f_s(T)$ and $f_f(T)$, are shown in Figures 2(a) and (b), respectively.

To map the data points between sub-sized and full-sized domains, we applied the following transformations for the temperature and absorbed energy.

*(a) Temperature transformation*: the temperature values for the sub-sized data points $T_s^i$ are shifted by the difference $\Delta T = \mathrm{DBTT}_f - \mathrm{DBTT}_s$, in order to align the DBTT of the sub-sized curve with that of the full-sized curve. The transformed temperatures $T_{\mathrm{tr}}^i$ are calculated as:

$$T_{\mathrm{tr}}^i = T_s^i + \Delta T = T_s^i + \left(\mathrm{DBTT}_f - \mathrm{DBTT}_s\right) \quad \text{for } i = 1,2,\dots,N \tag{5}$$

This transformation is illustrated in Figure 2(c), where the DBTT of shifted sub-sized data becomes 166 °C, compared to the original sub-sized DBTT of 203 °C. The approach resembles analytical correlation methods that apply a constant temperature shift; however, in this case, the shift is specimen-specific and based on experimentally determined DBTT values, rather than a generalized empirical or analytical value.

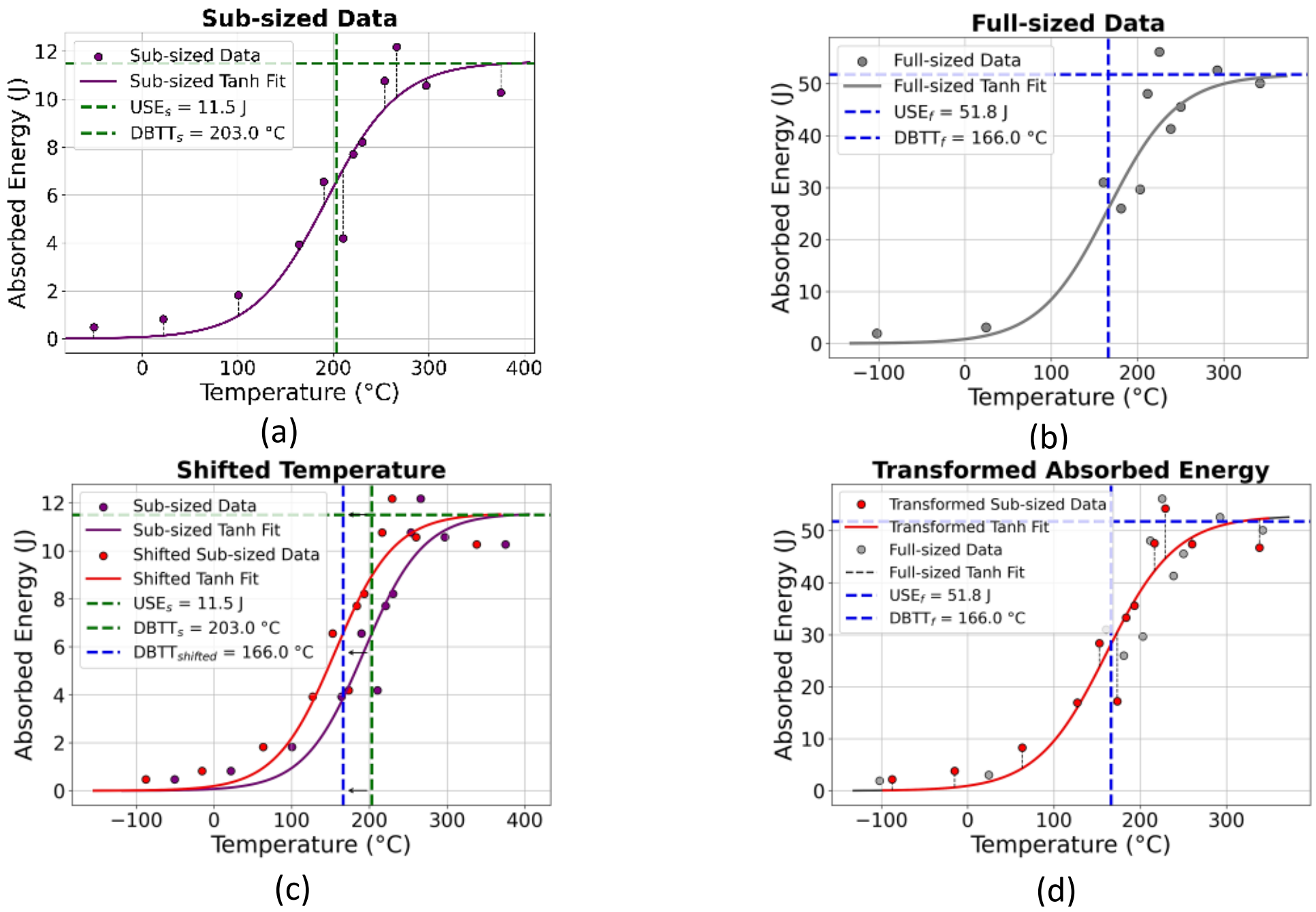

**Figure 2.** (a) Experimental data points for a sub-sized specimen set. (b) Experimental data points for the corresponding full-sized specimen set. (c) Sub-sized data points shift to align the DBTT (203 °C) with the full-sized DBTT (166 °C). (d) Sub-sized data points are vertically transformed to align the USE (11.5 J) with the full-sized USE (51.8 J). Experimental full-sized data points are also shown with gray circles for comparision.

*(b) Absorbed energy transformation*: to align the absorbed energy values with the full-sized specimen set, we applied a scaling transformation based on the residual deviations from the fitted sub-sized curve. Specifically, for each sub-sized data point, the residual $\Delta E^i$ is computed as the difference between the measured absorbed energy $E_s^i$ and the value predicted by the fitted model $f_s$ at temperature $T_s^i$:

$$\Delta E^i = E_s^i - f_s\left(T_s^i\right) \quad \text{for } i = 1,2,\dots,N \tag{6}$$

The residuals capture the variability of the Charpy data around the fitted curve and are indicated by dashed lines in Figure 2(a). To scale the energy deviations to the full-sized domain, the residuals are multiplied by the ratio of the USE of full-sized and sub-sized specimens:

$$\Delta E_{sc}^i = \Delta E^i \frac{\mathrm{USE}_f}{\mathrm{USE}_s} = \left(E_s^i - f_s\left(T_s^i\right)\right) \frac{\mathrm{USE}_f}{\mathrm{USE}_s} \quad \text{for } i = 1,2,\dots,N \tag{7}$$

To obtain the transformed absorbed energies in the full-sized domain $E_{tr}^{i}$, the scaled residuals $\Delta E_{sc}^{i}$ are added to the full-sized tanh model at temperatures adjusted for the DBTT shift $T_{\mathrm{tr}}^{i}$:

$$E_{tr}^{i} = f_f\left(T_{\mathrm{tr}}^{i}\right) + \left(E_s^i - f_s\left(T_s^i\right)\right)\frac{\mathrm{USE}_f}{\mathrm{USE}_s} \quad \text{for } i = 1,2,\dots,N \tag{8}$$

The results of the transformation are shown in Figure 2(d), where the scaled residuals are again depicted with dashed lines, and the full-sized experimental data are overlaid for comparison. When substituting the tanh model for $f_f$ in equation (8), the transformed absorbed energies take the following form:

$$E_{tr}^{i} = \frac{\mathrm{USE}_f}{2}\left(1 + \tanh\frac{T_{\mathrm{tr}}^{i} - \mathrm{DBTT}_f}{k_f}\right) + \left(E_s^i - f_s\left(T_s^i\right)\right)\frac{\mathrm{USE}_f}{\mathrm{USE}_s} \quad \text{for } i = 1,2,\dots,N \tag{9}$$

The rationale for incorporating the residuals $\Delta E^{i}$ is to preserve the relative variability of the sub-sized data with respect to the fitted sub-sized tanh curve, thereby ensuring that fine-scale fluctuations are retained as the energies are projected onto the full-sized tanh curve. The use of the slope parameter of the full-sized tanh model $k_f$ in equation (9) ensures that the transition sharpness of the resulting transformed data matches that of the actual full-sized Charpy curve. The objective of the proposed transformation is to capture both the baseline shape of the full-sized response and the specimen-specific scatter exhibited by the sub-sized data, thereby creating a distribution of transformed points that closely resembles that of the true full-sized data. For comparison, Figure 3 presents transformed and actual full-sized data for three additional representative specimens that supplement the information for the specimen set presented in Figure 2.

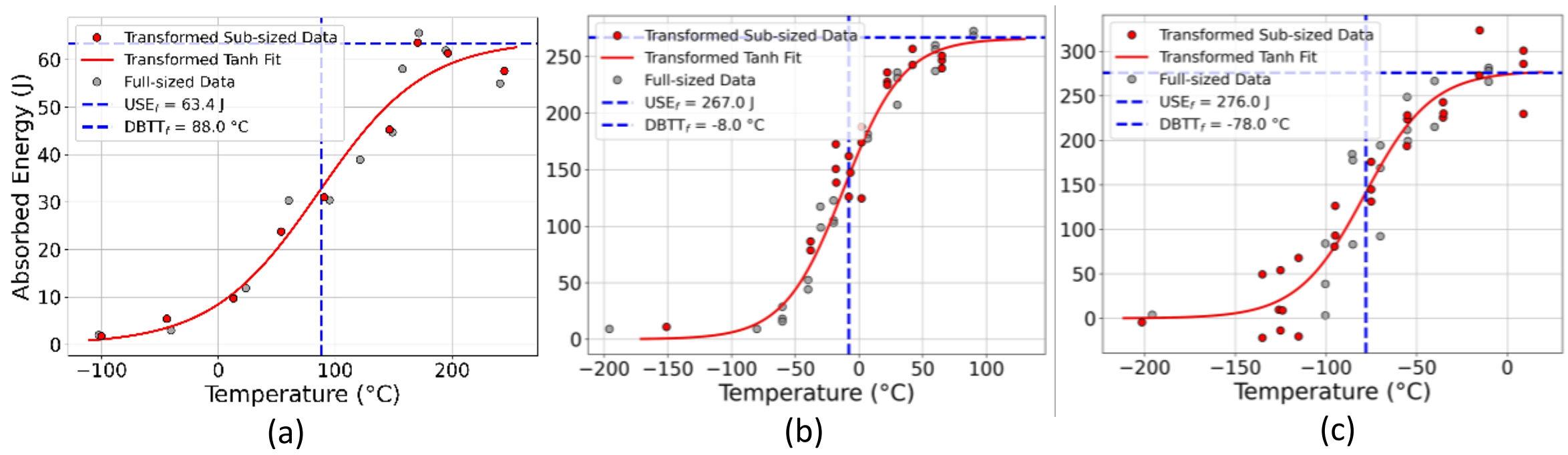

**Figure 3.** Transformed sub-sized data points (red circles) and experimental full-sized data points (gray circles) for three specimen sets.

### *3.2.2. Machine Learning Modeling*

The proposed ML modeling framework is illustrated in Figure 4. The model takes as input a collection of parameters that describe the material composition, manufacturing and treatment information, irradiation conditions, specimen geometry, and test parameters. Importantly, each data record includes temperature and absorbed energy from one Charpy sub-sized impact test. In total, there are 37 input parameters that include:

- Chemical composition (weight percent): C, Si, Mn, P, S, Ni, Cr, Mo, Fe, Cu, V.
- Manufacturing and heat treatment: manufacturer, pre-cracked specimen (yes/no), heat treatment type, final heat treatment temperature (°C), final heat treatment time (hours), cooling method, cooling rate (°C/min), microstructure, grain size (µm).
- Irradiation conditions: irradiated specimen (yes/no), neutron fluence (n/cm$^2$), irradiation temperature (°C), irradiation duration (hours).
- Specimen geometry: thickness (mm), width (mm), length (mm), span (mm), notch depth (mm), notch angle (deg), notch root radius (mm), specimen size (sub/full).

- Testing constraints: test standard, specimen orientation, product form (base metal, HAZ, or weld).
- Sub-sized Charpy data point: sub-sized test temperature (°C) $T_s$, sub-sized absorbed energy (J) $E_s$.

All 37 input parameters were retained in the modeling framework, because each corresponds to a known factor that can influence Charpy impact behavior. The parameters carry physically meaningful information, and collectively characterize the conditions under which Charpy responses are obtained. In exploratory trials, we evaluated reduced parameter sets, by removing subsets of compositional, geometric, or process-related variables, which resulted in decreased predictive performance. For this reason, no dimensionality reduction or parameter elimination was applied in the final workflow.

The prediction task is formulated as a multivariate regression problem. The model is trained to predict the transformed temperature (°C) and absorbed energy (J) corresponding to full-sized specimens, based on sub-sized Charpy test data and shared treatment and testing conditions. Specifically, the model learns the following mapping:

$$\begin{bmatrix} T_{tr} \\ E_{tr} \end{bmatrix} = f(x_1, x_2, x_3, \ldots, T_s, E_s) \tag{10}$$

where $T_{tr}$ and $E_{tr}$ are the transformed (full-sized domain) temperature and absorbed energy for a Charpy test, $f$ denotes the function learned by the model, $T_s$ and $E_s$ are sub-sized temperature and absorbed energy, and $x_1, x_2, x_3, \ldots$. denote other input parameters related to composition, treatment, geometry, and test parameters. This formulation enables the model to learn a mapping that accounts for temperature shifts and energy variability in Charpy impact behavior across size scales. Since each point on a Charpy impact curve is governed by the coupled relationship between temperature and absorbed energy, these two transformed quantities are predicted jointly, whereas the multivariate regression framework preserves this dependence, and leads to more stable and physically meaningful predictions.

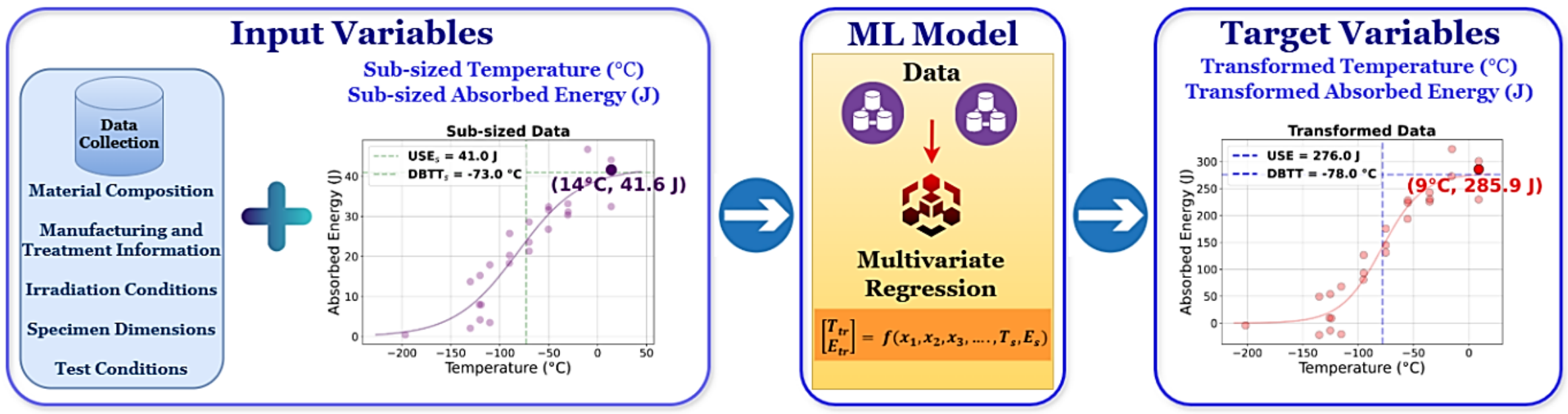

**Figure 4.** Workflow of ML model training. The model uses 37 input parameters related to material composition, treatment, test conditions, and sub-sized temperature and absorbed energy (e.g., 14 °C and 41.6 J in the figure). Multivariate regression is used to jointly predict two target variables: transformed temperature and transformed absorbed energy (e.g., 9°C and 285.9 J in the figure).

Once the ML model is trained, it can be used to infer the corresponding full-sized Charpy impact response for a given set of sub-sized Charpy test data. By fitting a tanh function to the predicted data points, estimates of the correlated USE and DBTT can be obtained. Importantly, the inference process does not require access to Charpy impact data from full-sized specimens. However, the reliability of the predicted correlated Charpy properties depends on whether the ML model has encountered specimens with similar processing, composition, geometry, and testing conditions during training.

# 4. Results

For a comparative analysis of the methods used for correlating Charpy impact properties, we employed standard performance metrics for continuous variables. The metrics include the coefficient of determination ($R^2$), the Pearson correlation coefficient ($r$), and root-mean-squared error (RMSE). A model is considered to perform better when it yields higher values of $R^2$ and $\boldsymbol{r}$, and a lower value of RMSE.

## 4.1. Correlation of Charpy Impact Properties Using Analytical Methods

The performance metric values for the analytical methods for correlating USE are presented in Table 4. For all specimens, we extracted both sub-sized and full-sized USE values from the pertinent publications, which allows utilizing the full-sized USE data as the reference for evaluating the correlated values produced by each method.

For the approaches that rely on empirical estimation of parameters, such as Uehira et al. [8] and Wallin et al. [37], the calculations of the correlated USE values can be sensitive to how the available data is used for calibration. Specifically, for the method proposed by Uehira et al. [8], the empirical equation presented in Table 1 can be applied by fitting a least-squares regression model to all available data points, resulting in the model: $0.411 \cdot (Bb)^{1.38}$, with parameters $m = 0.411$ and $n = 1.38$. However, this model yields relatively poor agreement with the full-sized USE values, achieving an $R^2$ value of 0.488, as shown in Table 4. This is largely due to the wide variability in the full-sized USE values, which range from 51.8 J to 278 J (see Table A1 in Supplementary Material). The observed variability is primarily attributed to differences in specimen preparation and treatment, which significantly influence Charpy impact behavior. An alternative approach involves applying the linear regression model to data collected from each publication separately, under the assumption that specimens within a single study underwent consistent treatment and processing. For example, the full-sized USE values from the study by Kim et al. [40] vary from 238 J to 278 J, which is substantially less than the range observed across all studies. Using this strategy, the parameters $m$ and $n$ can be estimated for the data from each reference individually, yielding the following models: $0.287 \cdot (Bb)^{1.19}$ [5], $0.32 \cdot (Bb)^{1.44}$ [40], $0.393 \cdot (Bb)^{1.39}$ [41], and $0.419 \cdot (Bb)^{1.42}$ [42]. This calibration strategy improves the correlation, achieving $R^2 = 0.824$. However, these models are specific to the treatment conditions of each dataset, and may not generalize to specimens subjected to different processing techniques.

Analogously, the approach by Wallin et al. [37], which is also based on an empirical relationship, can produce varying results depending on the used parameter estimation strategy. One strategy involves assigning different values to the exponent $m$ in the equation presented in Table 1, depending on specimen geometry. Similar to the approach employed by the authors [37], by fitting a linear regression model to the normalized USE values for sub-sized specimens $\mathrm{USE}_s/(B_s b_s)$ and full-sized specimens $\mathrm{USE}_f/(B_f b_f)$, the values of the exponent $m$ are as follows: 0.951 (5 mm thickness), 1.08 (3.33 mm thickness), 0.74 (2.5 mm thickness), and 0.929 (1.5 mm thickness). Incorporating these thickness-specific $m$ values into the equation for USE from Table 1 improved the quality of the correlation, achieving the highest performance among the tested models with $R^2 = 0.859$. An alternative method proposed by Wallin et al. [37] involves normalizing the absorbed energy of all sub-sized specimens to full-sized equivalents, followed by fitting a tanh function to estimate the correlated USE. The authors introduced the following approximation formula for the correlated absorbed energy values $\mathrm{E}_{cor}$:

$$\mathrm{E}_{cor} \approx \mathrm{E}_s \cdot \frac{b_f}{b_s} + \left(\frac{\mathrm{E}_s}{\mathrm{E}_f}\right)^4 \left(\mathrm{USE}_f - \mathrm{USE}_s \frac{b_f}{b_s}\right) \tag{11}$$

This approach achieved a strong correlation with $R^2 = 0.802$, albeit lower than the correlation obtained using the direct empirical relationship listed in the last row of Table 1.

We refer to the approach by Wallin et al. [37] in the last entry of Table 2 as indirect estimation of correlated USE, since it first uses equation (11) to correlate absorbed energies for full-sized specimens based on measured absorbed energies for sub-sized specimens, followed by fitting ductile-to-brittle transition curves to estimate USE values. Conversely, the remaining analytical methods presented in Table 1 apply direct USE estimation, since they predict the full-sized USE as a direct function of the measured sub-sized USE.

The slightly lower performance of the indirect normalization approach in equation (11) may be attributed to the absence of case-specific parameter calibration. Unlike the case-dependent direct formulations, which incorporate optimized exponent values calibrated to specific specimen geometries or datasets, the normalization-based method applies a more generalized transformation of absorbed energy prior to curve fitting. While this reduces sensitivity to dataset-specific parameter tuning, it may also limit the generalizability within a heterogeneous dataset.

Consequently, the empirical approaches developed by Wallin et al. [37] and Uehira et al. [8] demonstrated the best performance, particularly when the model parameters are calibrated to account for specimen treatment or geometry. Additionally, the correlation methods proposed by Lucas et al. [16,17] and Corwin et al. [18] also showed good agreement with the actual full-sized USE values.

**Table 4**. Performance metrics of analytical models for USE correlation.
[1]Uehira correlation method applied to data points from each reference individually.
[2]Wallin correlation method used for indirect estimation of correlated USE from fitted tanh curves.

| Correlation Method | $R^2$ ↑ | $r$ ↑ | RMSE ↓ |
|---|---|---|---|
| **Lucas et al. [16,17]** | 0.723 | 0.893 | 42.2 |
| **Corwin et al. [18]** | 0.720 | 0.902 | 42.4 |
| **Louden et al. [19]** | -0.572 | 0.805 | 100.6 |
| **Kayano et al. [20]** | 0.489 | 0.888 | 57.4 |
| **Schubert et al. [5]** | -0.610 | 0.753 | 101.8 |
| **Uehira et al. [8]** | 0.488 | 0.880 | 57.4 |
| **Uehira et al. [8][1]** | 0.824 | 0.939 | 33.7 |
| **Wallin et al. [37]** | **0.859** | **0.944** | **30.2** |
| **Wallin et al. [37][2]** | 0.802 | 0.923 | 35.7 |

The performance metric values for the analytical methods for correlating DBTT are presented in Table 5. Similar to the USE estimation, the empirical method proposed by Uehira et al. [8] can be applied to the available data to calibrate the model parameters. For instance, when the relationships for DBTT correlations are derived individually based on data from each publication, the following models are obtained: $-11 \cdot \log_{10}(B_s K_{t,s}) - 158$ [5], $54 \cdot \log_{10}(B_s K_{t,s}) - 122$ [40], $109 \cdot \log_{10}(B_s K_{t,s}) - 188$ [41], and $111 \cdot \log_{10}(B_s K_{t,s}) - 201$ [42]. Among the evaluated methods, the approach proposed by Towers [24] achieved the best overall performance. Overall, DBTT correlation is more challenging than USE, as the full-sized DBTT temperatures vary from -78 °C to 160 °C (see table A1). Furthermore, while some sub-sized specimens exhibit a decrease in DBTT, the majority show an increase, adding to the complexity of accurately correlating DBTT values.

The negative $R^2$ observed for the Louden et al. [19] correlation indicates that the predicted values deviate more from the experimental full-sized DBTT values than a simple mean-value prediction. This behavior likely reflects the sensitivity of the empirical formulation to the statistical distribution and variability of the compiled dataset. The correlation in Louden et al. [19] was originally derived based on experiments with ferritic–martensitic steel HT-9, whereas the assumptions do not hold for the dataset used in our study. Also, some of the other analytical correlation methods were developed by using small datasets that focused on limited number of materials. The correlation proposed by Towers [24] appears comparatively robust for the present dataset, possibly because its formulation includes several different alloys for developing the correlation relationship, and it is a simple additive method that apply a shift in the transition temperature without introducing additional parameters that need to be calibrated.

**Table 5**. Performance metrics of analytical models for DBTT correlation.
[1]Uehira correlation method applied to data points from each reference individually.

| Correlation Method | $R^2$ ↑ | *r* ↑ | RMSE ↓ |
|---|---|---|---|
| **Towers [24]** | **0.661** | **0.899** | **35.5** |
| **Louden et al. [19]** | -4.944 | 0.801 | 148.4 |
| **Kayano et al. [20]** | 0.493 | 0.884 | 43.3 |
| **Sokolov et al. [6]** | 0.435 | 0.881 | 45.8 |
| **Lucon et al. [25]** | 0.485 | 0.884 | 43.7 |
| **Uehira et al. [8]** | 0.045 | 0.337 | 59.5 |
| **Uehira et al. [8][1]** | 0.498 | 0.876 | 43.1 |
| **Wallin et al. [37]** | 0.584 | 0.879 | 39.3 |

### 4.2. Correlation of Charpy Impact Properties Using Machine Learning

The ML models were trained exclusively on Charpy impact data for sub-sized specimens, or in other words, data from full-sized specimens were not used for model training. On the other hand, full-sized specimen data were used for model assessment and for sub-sized data transformation. Specifically, the ML models did not receive any full-sized measurements as input parameters for model training, and instead, full-sized specimen information was used exclusively to generate the target values that the model is trained to approximate. The dataset contained 389 test records obtained from 27 sub-sized specimen sets. To evaluate model performance, we employed a specimen-based cross-validation strategy, where in each of five iterations, data from 5 or 6 specimen sets were set aside for testing, while the remaining data from 21 or 22 specimen sets were used for training. The corresponding test partitions contained 77, 87, 65, 78, and 55 individual temperature-energy datapoints. This process was repeated five times, resulting in five independently trained models. The reported performance metrics represent the average value and the standard deviation across the five runs. Such specimen-level data partitioning strategy ensures that each of the 27 specimen configurations is included in the test set for one of the five runs, and that the models were not evaluated on specimens seen during training, preventing information leakage (i.e., unintended transfer of information from the test data into the training process).

The experimental evaluation involved seven ML models, comprising both traditional and modern approaches. The models include Random Forest, Gaussian Process Regression (GPR), Gradient Boosting, Light Gradient Boosting Machine (LightGBM), Categorical Boosting (CatBoost), Extreme Gradient Boosting (XGBoost), and Neural Networks (NNs) with three fully connected hidden layers having 128, 64, and 32 neurons, respectively. The models were implemented using the scikit-learn Python library. We conducted an extensive grid search for hyperparameter optimization for all models. Performance metric values for each model are presented in Table 6. Overall, the best results were achieved by ensemble tree-based methods, which include Gradient Boosting, CatBoost,

XGBoost, and LightGBM. Examples of predicted temperature-absorbed energy data points and the target data points for three specimens are presented in Figure 5.

For the test specimens, the values of USE and DBTT were determined by fitting a tanh function to the temperature-absorbed energy data. To fit the tanh function to the experimental data, we employed a nonlinear regression function from the SciPy library. The model estimated values were compared to the benchmark USE and DBTT values from the corresponding full-sized specimens, following the same procedure used for the analytical correlation methods. The resulting performance metrics are reported in Table 7. In comparison to the best-performing analytical models (Tables 4 and 5) which achieved $R^2 = 0.859$ for USE and $R^2 = 0.661$ for DBTT, the correlated values by the tree-ensemble models (e.g., for the Gradient Boosting model $R^2 = 0.942$ for USE and $R^2 = 0.830$ for DBTT) align more closely with the full-sized USE and DBTT values and produce higher values for the performance metrics. Among the evaluated ML models, Gradient Boosting and CatBoost achieved the highest predictive accuracy for absorbed energy (USE), with $R^2$ values of 0.846 and 0.853, respectively. For DBTT prediction, Gradient Boosting demonstrated the strongest performance ( $R^2$ = 0.941 and RMSE = 19.30), indicating superior capability in capturing the transition behavior. Although CatBoost slightly outperformed other models in USE for the $R^2$ metric, Gradient Boosting provided the most balanced and consistent performance across both USE and DBTT metrics. Therefore, considering predictive accuracy, stability across cross-validation iterations, and overall robustness, the Gradient Boosting model is recommended as the preferred ML framework for the proposed correlation approach.

**Table 6**. Performance metrics of ML Models for Absorbed Energy and Temperature. The values show the average and standard deviation over five training and testing iterations.

| ML Models | Absorbed Energy | | | Temperature | | |
|---|---|---|---|---|---|---|
| | $R^2$ ↑ | $r$ ↑ | RMSE ↓ | $R^2$ ↑ | $r$ ↑ | RMSE ↓ |
| **XGBoost** | $0.801_{\pm 0.10}$ | $0.901_{\pm 0.05}$ | $41.20_{\pm 13.2}$ | $0.894_{\pm 0.07}$ | $0.964_{\pm 0.03}$ | $22.30_{\pm 4.9}$ |
| **LightGBM** | $0.814_{\pm 0.07}$ | $0.904_{\pm 0.03}$ | $41.77_{\pm 10.8}$ | $0.877_{\pm 0.07}$ | $0.937_{\pm 0.01}$ | $27.81_{\pm 7.0}$ |
| **Random Forest** | $0.750_{\pm 0.09}$ | $0.867_{\pm 0.04}$ | $48.49_{\pm 12.2}$ | $0.914_{\pm 0.07}$ | $0.960_{\pm 0.01}$ | $23.19_{\pm 3.7}$ |
| **GPR** | $0.741_{\pm 0.06}$ | $0.861_{\pm 0.02}$ | $49.32_{\pm 4.8}$ | $0.893_{\pm 0.08}$ | $0.948_{\pm 0.04}$ | $25.967_{\pm 7.0}$ |
| **Neural Networks** | $0.722_{\pm 0.23}$ | $0.879_{\pm 0.06}$ | $51.13_{\pm 19.3}$ | $0.876_{\pm 0.30}$ | $0.941_{\pm 0.01}$ | $27.89_{\pm 13.0}$ |
| **Gradient Boosting** | $0.846_{\pm 0.07}$ | $0.920_{\pm 0.03}$ | $38.02_{\pm 10.4}$ | $\mathbf{0.941_{\pm 0.08}}$ | $\mathbf{0.971_{\pm 0.02}}$ | $\mathbf{19.30_{\pm 5.2}}$ |
| **CatBoost** | $\mathbf{0.853_{\pm 0.05}}$ | $\mathbf{0.925_{\pm 0.02}}$ | $\mathbf{37.13_{\pm 8.7}}$ | $0.926_{\pm 0.07}$ | $0.964_{\pm 0.01}$ | $21.53_{\pm 5.3}$ |

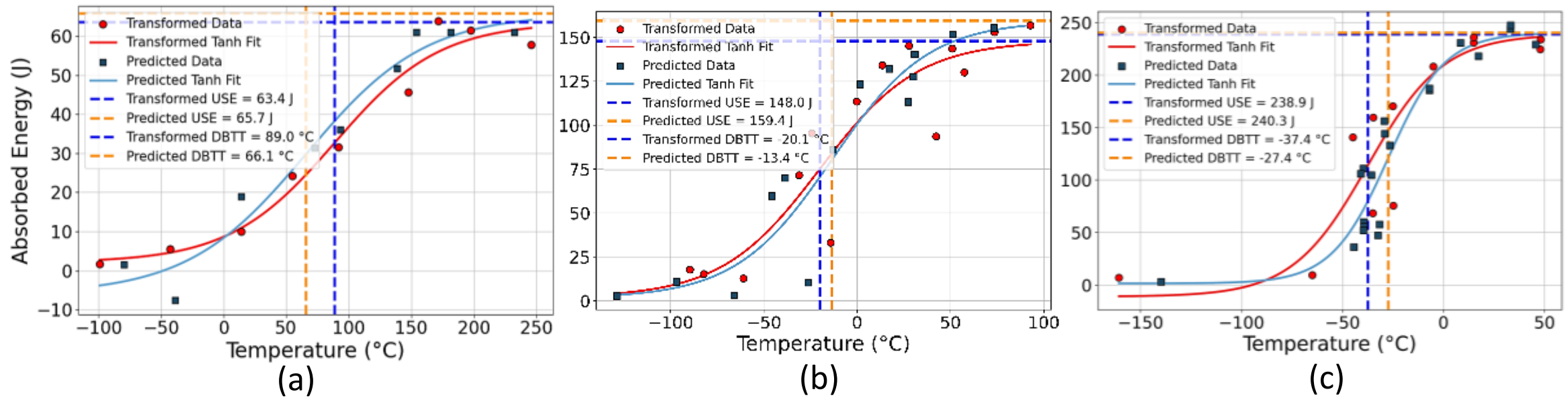

**Figure 5.** Predicted (blue squares) and transformed (red circles) temperatures and absorbed energies for three specimen sets using the LightGBM model.

A statistical significance analysis was conducted using a Wilcoxon signed-rank test to compare paired samples consisting of correlated USE and DBTT values and the corresponding experimental ground-truth full-size specimen values for each ML and analytical correlation method. The null hypothesis states that the median difference between the predicted correlated values and the ground-truth full-size specimen values is zero. The results from the Wilcoxon test for the top four ML and analytical correlation methods are presented in Table 8. For the ML methods, both for USE and DBTT, the obtained *p*-values are well above the adopted significance level of 0.05. Therefore, we fail to reject the null hypothesis and conclude that the difference between the correlated USE and DBTT values and the ground-truth values is not statistically significant at the adopted significance level. For the analytical/empirical correlation methods, several methods produce *p*-values below 0.05, indicating statistically significant systematic bias, whereas other methods do not show statistically significant differences. It should be noted that a high *p*-value for the Wilcoxon signed-rank test indicates that no statistically detectable systematic directional bias is present. The variation in *p*-values across ML methods is due to differences in the distribution of residuals across models. The superior predictive accuracy of ML methods relative to analytical/empirical methods is supported by the $R^2$ and RMSE comparisons presented in Table 7.

**Table 7**. Performance metrics for ML-predicted correlations of USE and DBTT.

| ML Models | USE | | | DBTT | | |
|---|---|---|---|---|---|---|
| | $R^2$ ↑ | *r* ↑ | RMSE ↓ | $R^2$ ↑ | *r* ↑ | RMSE ↓ |
| **XGBoost** | 0.913 | 0.963 | 24.1 | 0.814 | 0.938 | 27.2 |
| **LightGBM** | 0.901 | 0.965 | 24.2 | **0.892** | **0.948** | **17.8** |
| **Random Forest** | 0.894 | 0.961 | 26.3 | 0.776 | 0.886 | 29.2 |
| **GPR** | 0.831 | 0.914 | 33.3 | 0.786 | 0.894 | 28.6 |
| **Neural Networks** | 0.837 | 0.950 | 32.7 | 0.798 | 0.896 | 27.8 |
| **Gradient Boosting** | **0.942** | **0.974** | **19.7** | 0.830 | 0.913 | 27.0 |
| **CatBoost** | 0.937 | 0.970 | 20.2 | 0.838 | 0.920 | 24.9 |

**Table 8:** Statistical significance results from Wilcoxon signed-rank test for the top performing ML and analytical/empirical correlation methods for USE and DBTT.

| Correlation Method | USE *p*-value | Statistically Different | Correlation Method | DBTT *p*-value | Statistically Different |
|---|---|---|---|---|---|
| **Gradient Boosting** | **0.988** | Yes | **LightGBM** | **0.854** | Yes |
| **CatBoost** | 0.803 | Yes | **XGBoost** | 0.722 | Yes |
| **XGBoost** | 0.672 | Yes | **CatBoost** | 0.712 | Yes |
| **LightGBM** | 0.063 | Yes | **Gradient Boosting** | 0.675 | Yes |
| **Wallin et al. [37]** | 0.708 | Yes | **Towers [24]** | 0.249 | Yes |
| **Uehira et al. [8][1]** | 0.332 | Yes | **Uehira et al. [8][1]** | 0.014 | No |
| **Lucas et al. [16,17]** | 0.029 | No | **Wallin et al. [37]** | 0.009 | No |
| **Corwin et al. [18]** | 0.004 | No | **Kayano et al. [20]** | 0.001 | No |

### 4.3. Parameter Importance Analysis

To interpret the developed ML models and identify which parameters most significantly influence their predictions, we conducted a post-hoc explainability analysis using SHAP (SHapley Additive exPlanations) [44]. SHAP was developed in cooperative game theory, where Shapley values are a solution concept to fairly distribute the total payout among players in a cooperative game based on their individual contributions across all possible coalitions. In the context of ML explainability, each input parameter is treated as a player, and its Shapley value represents the average marginal contribution of that parameter to the model's prediction across all possible subsets of parameters.

This study employs Shapley values to quantify the marginal contribution of each input parameter to the output of a model, providing insights into the relative importance of material properties, treatment conditions, specimen geometry, test conditions, and other input parameters in predicting Charpy impact properties.

The most important input parameters for the trained tree-ensemble models are shown in Figure 6(a) and (b), corresponding to predictions of the transformed absorbed energy $E_{tr}$ and temperature $T_{tr}$, respectively. The sub-sized absorbed energy $E_s$ and temperature $T_s$ are the most important input parameters for predicting the correlated full-sized values. Additional influential parameters for energy prediction include specimen dimension parameters, such as width, thickness, and notch depth. Also, the carbon content (C, wt. %) and product form, i.e., whether the specimen was taken from the base material, weld metal, or heat-affected zone, affect the absorbed energy. For the temperature shift prediction analysis shown in Figure 6(b), specimen dimension parameters are also dominant, along with smaller contributions by the elemental content of manganese (Mn, wt. %) and silicon (Si, wt. %).

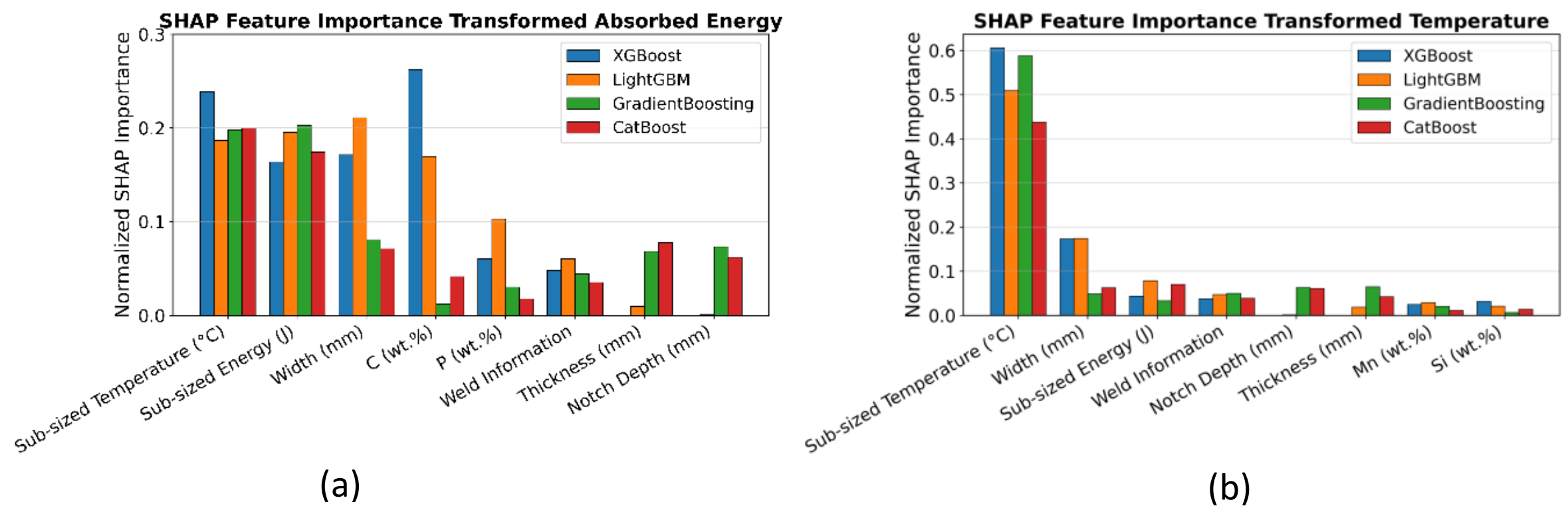

**Figure 6.** SHAP parameter importance values for the trained tree-ensemble models: (a) transformed absorbed energy, and (b) transformed temperature.

An additional SHAP analysis was performed using models trained solely on full-sized Charpy test data for SA533B steel. A total of 869 Charpy test records were used to train the models to predict absorbed energy, with the SHAP analysis presented in Figure 7(a). These models were trained and analyzed for the purpose of eliminating the specimen dimensional effects, and focus on other parameters affecting Charpy impact properties. Since all specimens in this dataset are full-sized, specimen dimensions do not contribute to the predictive performance. Besides test temperature as the most important parameter with SHAP values above 0.35 for all models, chemical composition and irradiation conditions significantly influenced the predictions, with the contents of Fe, C, S, and Ni, as well as the irradiation dose, being the most impactful parameters across all models. Furthermore, a separate collection of ML models was trained using a combined dataset of full-sized and sub-sized Charpy specimens to predict absorbed energy. The corresponding SHAP analysis is shown in Figure 7(b). In this case, the models relied heavily on the specimen size parameter, which is a categorical parameter, indicating whether a specimen is sub-sized or full-sized. Specimen dimensions also had a strong influence on the predictions, and among the compositional parameters S, C, and Si emerged as key contributors to models' output.

Conclusively, the SHAP analysis confirms that the most influential ML parameters align with well-established factors affecting Charpy impact behavior, such as test temperature, specimen geometry, treatment processes, and alloy composition. Specimen location, such as weld location information and specimen orientation, can dictate the microstructure and morphology of the specimen. The treatment conditions can heavily influence microstructural changes and morphology of the material. Both specimen location and treatment conditions can introduce complex dependencies that influence mechanical behavior, which plays an essential role in Charpy impact test response. The importance of specimen dimensions and geometry is expected and reflects the geometry-dependent nature of fracture mechanics and energy absorption during impact. Additionally, numerous studies have reported that neutron irradiation embrittlement reduces the material's ability to absorb energy during fracture and can lead to a shift in DBTT [45–47]. This degradation caused by neutron irradiation is attributed to factors such as displacement damage and radiation-induced segregation, which reduce the ductility and toughness of the material [48].

Alloy composition parameters also affect the materials' mechanical response. Ahmedabadi [27] compiled a comprehensive dataset of Charpy tests on plain carbon steel alloys with varying carbon content, and reported a trend of decreasing absorbed energy and increasing DBTT with increasing carbon content. Even small variations in carbon content can lead to significant changes in the microstructure, which can further impede dislocation movements, and thereby increase strength, but sacrifice ductility in the process. Other alloying elements, such as sulfur, can form sulfide inclusions at the grain boundaries leading to embrittlement and a reduction in absorbed energy. This effect has been well studied and is especially important in welding applications, where sulfur-induced cracking can be pronounced [49–51]. Nickel is widely used in stainless steels to stabilize austenitic phase and improve ductility and toughness. However, in SA533B low-alloy steels, the specified nickel content is relatively low, thus minimizing the effects of the austenitic phase. Moreover, nickel can form strengthening precipitates, which, while improving strength, often negatively affect Charpy impact toughness [52].

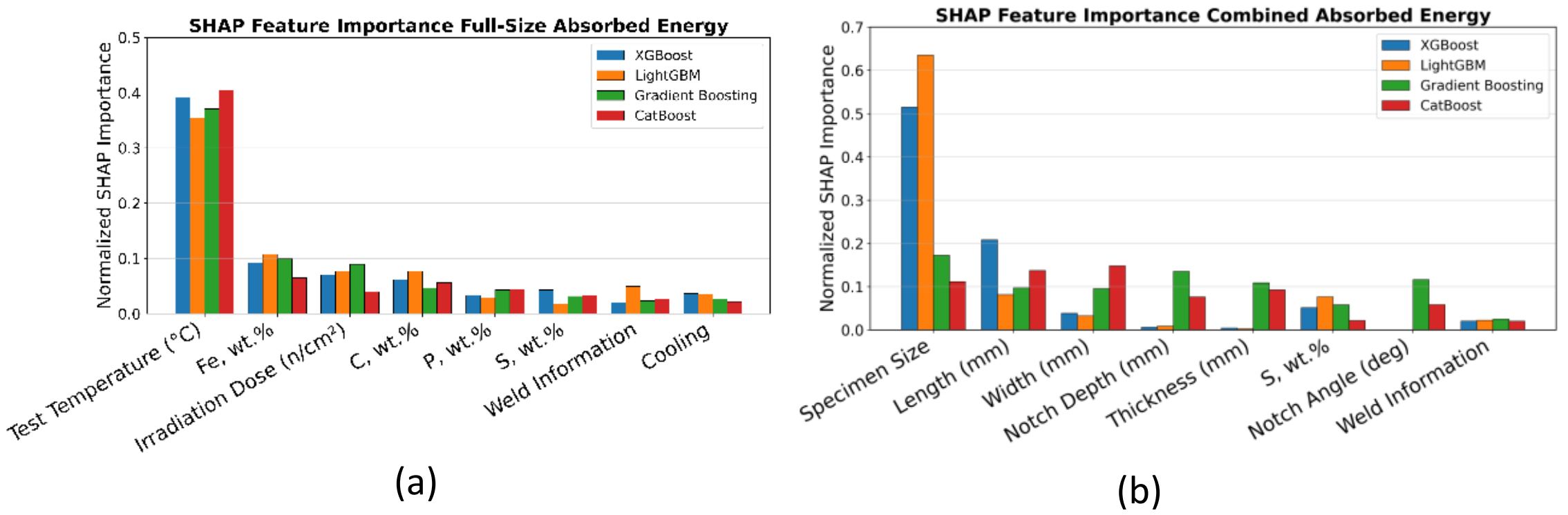

**Figure 7.** SHAP parameter importance values for absorbed energy using (a) models trained on full-sized Charpy data, and (b) models trained on combined full-sized and sub-sized Charpy data.

### 4.4. Uncertainty Quantification

To assess the reliability of the correlated Charpy impact properties and support informed decision-making in practical applications, uncertainty quantification (UQ) is of utmost importance. Prior works have investigated UQ in Charpy impact testing based on sources of uncertainties originating from the various components in impact test machines and associated equipment [53−56], measurement instruments [57], as well as empirically estimated uncertainty in Charpy data based on numerical simulation using Monte Carlo sampling [58,59]. In this section, we present three complementary strategies for UQ for the USE and DBTT parameters obtained by ML-based Charpy data correlation.

Strategy 1: This approach uses the covariance matrix of the fitted tanh function to full-sized specimen Charpy data to estimate confidence intervals for USE and DBTT [60]. For experimental data points $(T_s, E_s)$, equation (4) is used to fit the hyperbolic tangent model $f(T)$ where $\theta$ are the fitted parameters obtained via nonlinear least squares. For the parameter covariance matrix $\Sigma_\theta$, the standard error of the *j*-th parameter is:

$$\sigma_{\theta_j} = \sqrt{(\Sigma_\theta)_{jj}},\ for\ j \in \text{USE, DBTT}, k \tag{12}$$

The 95 % confidence intervals for USE and DBTT are estimated based on $\pm 2$ standard deviations from full-sized Charpy test data:

$$CI_{USE} = [USE_f - 2\sigma_{\mathrm{USE}}, USE_f + 2\sigma_{\mathrm{USE}}]; \quad CI_{DBTT} = [DBTT_f - 2\sigma_{\mathrm{DBTT}}, DBTT_f + 2\sigma_{\mathrm{DBTT}}] \quad (13)$$

Figure 8(a) shows the Charpy data and the corresponding tanh fit for a full-sized specimen set. Based on the variability in the experimental data, the 95 % confidence level for the estimated USE is $228 \text{ J} \pm 20.5 \text{ J}$, and for the estimated DBTT is $-33.7\ °\mathrm{C} \pm 4.9\ °\mathrm{C}$. This procedure is applied to all full-sized specimen sets to quantify the variability in USE and DBTT.

Next, we compute the percentage of predicted USE and DBTT values from ML and analytical models that fall within the confidence intervals. The calculated percentages serve as coverage metrics $c_{\mathrm{USE}}$ and $c_{\mathrm{DBTT}}$ for uncertainty estimation. For the $i$-th specimen set, the coverages are calculated as:

$$c_{\mathrm{USE}} = \frac{1}{N}\sum_{i=1}^{N} \mathbf{1}\left[USE_{cor}^{i} \in CI_{USE}^{i}\right]; \quad c_{\mathrm{DBTT}} = \frac{1}{N}\sum_{i=1}^{N} \mathbf{1}\left[DBTT_{cor}^{i} \in CI_{DBTT}^{i}\right] \quad (14)$$

where $USE_{cor}^{i}$ and $DBTT_{cor}^{i}$ denote the correlated values, and $\mathbf{1}$ is the indicator function.

The coverage values for USE and DBTT are summarized in Tables 9 and 10. Among the ML models, CatBoost achieved the highest USE coverage with 76.9 %, while LightGBM attained the highest DBTT coverage with 57.7 %. Among analytical/empirical models, the method by Uehira et al. [8] provided the highest USE coverage with 73.0 %, and Kayano et al. [20] achieved the highest DBTT coverage with 34.6 %.

Strategy 2: In this approach, the predicted data points by ML models are used to fit a tanh function, from which prediction intervals for USE and DBTT are estimated based on the covariance matrix of the fit. Specifically, for predicted data points $(\hat{T}_{tr}, \hat{E}_{tr})$, equation (4) is used to fit the hyperbolic tangent model $f(\hat{T}_{tr})$, and the standard errors of the parameters are estimated from the covariance matrix of the nonlinear least squares. The 95 % confidence intervals are calculated based on $\pm 2$ standard deviations from estimated USE and DBTT for predicted data:

$$CI_{USE} = [\widehat{USE}_{tr} - 2\sigma_{\mathrm{USE}}, \widehat{USE}_{tr} + 2\sigma_{\mathrm{USE}}]; \; CI_{DBTT} = [\widehat{DBTT}_{tr} - 2\sigma_{\mathrm{DBTT}}, \widehat{DBTT}_{tr} + 2\sigma_{\mathrm{DBTT}}] \quad (15)$$

An example is illustrated in Figure 8(b). Coverage is calculated as the percentage of ground-truth full-sized values falling within ±2 standard deviations around the predicted values using equation (14). The results shown in Table 11 indicate that this strategy yields higher coverage metrics, where for USE up to 96.2 % of full-sized values fall within the prediction intervals for ML models, while DBTT reaches 73.1 % coverage. This approach was not applied to analytical/empirical models, as they typically do not allow for uncertainty quantification on DBTT or USE, since they are deterministic models without stochasticity.

Strategy 3: This strategy extends Strategy 2, however, instead of relying on ML models that output single-point predictions, it utilizes models that output the variance around the predictions. Such ML models include GPR, Quantile Regression, Bayesian Neural Networks, Natural Gradient Boosting, Monte Carlo Dropout, bootstrapped ensembles, and others [61]. Additionally, tree-ensemble models can be used for uncertainty predictions using quantile regression to generate lower and upper prediction bounds.

The estimated variances in temperature and absorbed energy predictions are used as weights in the tanh fitting process via inverse variance weighting. The optimal parameters $\hat{\theta}$ are calculated by dividing the residuals between ML-predicted absorbed energy and the tanh model $f(T_{s,i}; \theta)$, divided by the predicted variance from the ML model $\sigma_{E,i}^{2}$:

$$\hat{\theta} = \underset{\theta}{\operatorname{argmin}} \sum_{i=1}^{n} \frac{\left(\hat{E}_{tr} - f(T_{s,i};\theta)\right)^2}{\sigma_{E,i}^2} \tag{16}$$

Prediction intervals for USE and DBTT are then derived from the covariance of the fit, and coverage is again calculated based on the fraction of ground-truth values that fall within the prediction intervals, similar to Strategy 2. The uncertainty regions for the predictions by the GPR model are shown with ellipses around the data points in Figure 8(c). The figure also shows the corresponding 95 % prediction intervals for USE and DBTT based on weighted tanh fit, which accounts for uncertainty in both the temperatures and absorbed energies. The coverage values for USE and DBTT are presented in Table 12.

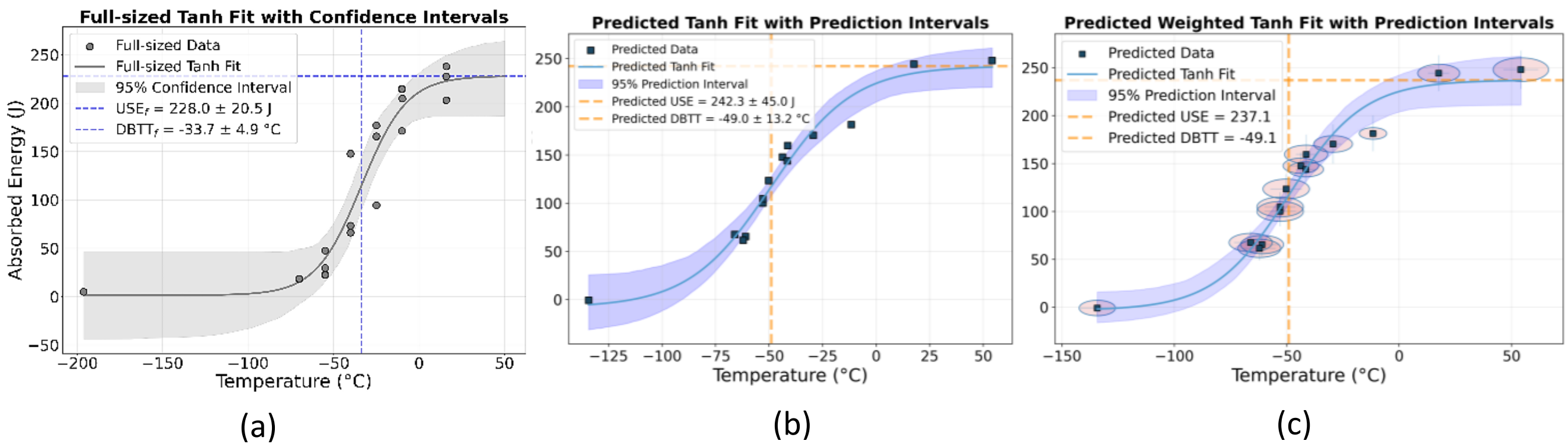

**Figure 8.** Uncertainty quantification using fitted tanh functions: (a) Strategy 1 - confidence intervals for USE and DBTT estimated from full-sized Charpy test data; (b) Strategy 2 - prediction intervals for USE and DBTT estimated from ML-predicted data points; (c) Strategy 3 - prediction intervals using ML-predicted data points with uncertainty weights shown as ellipses around the data points.

**Table 9:** Percentage of predicted USE and DBTT values from ML models within ±2 standard deviations of full-sized values.

| ML Models | USE Coverage | DBTT Coverage |
|---|---|---|
| **XGBoost** | 64.0 % | 20.0 % |
| **LightGBM** | 61.5 % | **57.7 %** |
| **CatBoost** | **76.9 %** | 50.0 % |
| **Gradient Boosting** | 69.6 % | 30.4 % |

**Table 10:** Percentage of predicted USE and DBTT from analytical correlation methods within ±2 standard deviations of full-sized values.

| Correlation Method | USE Coverage | Correlation Method | DBTT Coverage |
|---|---|---|---|
| **Lucas et al. [16,17]** | 38.4 % | **Towers [24]** | 15.3 % |
| **Corwin et al. [18]** | 46.1 % | **Louden et al. [19]** | 11.5 % |
| **Louden et al. [19]** | 15.3 % | **Kayano et al. [20]** | **34.6 %** |
| **Kayano et al. [20]** | 26.9 % | **Sokolov et al. [6]** | 23.0 % |
| **Schubert et al. [5]** | 3.8 % | **Lucon et al. [25]** | 26.9 % |
| **Uehira et al. [8]** | 30.7 % | **Uehira et al. [8]** | 15.3 % |
| **Uehira et al. [8][1]** | **73.0 %** | **Uehira et al. [8][1]** | 15.3 % |
| **Wallin et al. [37]** | 53.8 % | **Wallin et al. [37]** | 15.3 % |

**Table 11:** Percentage of full-sized experimental USE and DBTT within ±2 standard deviations of values derived from ML-based single-point predictions.

| ML Models | USE Coverage | DBTT Coverage |
|---|---|---|
| **XGBoost** | 84.6 % | 65.4 % |
| **LightGBM** | 80.8 % | **73.1 %** |
| **CatBoost** | 92.3 % | 69.2 % |
| **Gradient Boosting** | **96.2 %** | 80.8 % |

**Table 12:** Percentage of full-sized experimental USE and DBTT within ±2 standard deviations of values derived from ML-based variance around predictions.

| ML Models | USE Coverage | DBTT Coverage |
|---|---|---|
| **XGBoost** | 69.2 % | 73.1 % |
| **LightGBM** | 73.1 % | 42.3 % |
| **Gradient Boosting** | **92.3 %** | 69.2 % |
| **GPR** | 76.0 % | **80.0 %** |

A key challenge in validating UQ strategies for Charpy impact tests is the limited availability of repeated experiments performed under identical conditions in existing datasets. This limitation primarily affects the characterization and validation of aleatoric uncertainty, which reflects the inherent variability in Charpy impact test results arising from material heterogeneity and microstructural variations. Reliable estimation of the actual variability in USE and DBTT requires multiple tests conducted at the same temperature and under identical experimental conditions. Although in some cases, such as the example shown in Figure 8(a), researchers have carried out repeated tests at specific temperatures (e.g., three tests at 20 °C) with commendable experimental rigor, such instances remain exceptions rather than the norm. Consequently, rigorous validation of UQ approaches in Charpy impact testing remains challenging. A comprehensive uncertainty quantification framework should ideally distinguish between aleatoric uncertainty arising from experimental variability and epistemic uncertainty associated with limited data and model assumptions.

## 5. Discussion

The evaluation of the proposed ML-based methodology for correlating Charpy impact properties of sub-sized specimens indicates improved agreement with full-sized specimen results with respect to conventional analytical correlation methods. This improvement originates from the ability of ML models to capture arbitrary nonlinear relationships among Charpy properties without being limited to predefined functional forms such as linear or power-law relationships. Another advantage of ML approaches is the capacity to incorporate heterogeneous data types, including categorical parameters such as heat treatment, microstructural characteristics, specimen orientation, test conditions, and other factors that are difficult to incorporate into analytical models, but can have a significant impact on Charpy response. Moreover, explainable ML techniques can be applied to quantify parameter importance, offering insights into the underlying mechanisms that influence size effects in Charpy testing.

Furthermore, analytical methods that rely on empirically calibrated parameters (e.g., Uehira et al. [8] and Wallin et al. [37]) achieved improved performance when calibrated with specimens having specific processing or treatment conditions. However, such parameter tuning reduces generalizability when applied to new materials or specimen configurations. ML models, by contrast, benefit when trained on data with diverse processing or treatment conditions, which enhances their generalization ability.

Despite these advantages, the present study also has limitations. In particular, the experimental validation was performed only on SA533B steel, and the dataset included a relatively small number of 27 sub-sized specimen sets, with a total of 389 Charpy test records. Expanding the validation to larger datasets that include different materials and processing conditions will improve the applicability of the proposed framework. These limitations highlight the need for coordinated efforts among academic institutions and industry organizations to compile comprehensive databases and promote collaborative data-sharing initiatives.

# 6. Conclusion

This paper presents the first ML framework for correlating Charpy impact properties of sub-sized specimens with full-sized specimens of RPV steels. To this end, we collected Charpy test data from sub-sized specimens and corresponding full-sized data from materials with comparable processing and treatment records. The proposed approach maps absorbed energy-temperature data points from sub-sized specimens to the full-sized domain by applying a data-driven temperature shift combined with a residual-based energy transformation, aiming to approximate the full-sized data distribution across the entire ductile-to-brittle transition region and in upper shelf. Validation using a dataset of 389 Charpy test results for the SA533B steel showed that the ML-based approach achieved improved accuracy and consistency compared to established analytical methods for estimating USE and DBTT parameters. Using the trained ML models does not require access to full-sized data, and for a given set of Charpy impact test results from sub-sized specimens, the models can directly output the correlated energy-temperature data of full-sized specimens. This approach allows for a more reliable use of sub-sized Charpy specimens in nuclear surveillance programs, structural integrity assessments, and material development workflows, as well as broadly addresses size effects in restrictive experimental settings with limited material availability or complex component geometries.

## Funding

This work was supported by the U.S. Department of Energy, Office of Nuclear Energy under DOE Idaho Operations Office Contract DE-AC07-05ID14517 and LDRD project of 24A1081-149FP. The authors also acknowledge the support of DOE Advanced Fuel Campaign on experimental data collection. Accordingly, the publisher, by accepting the article for publication, acknowledges that the U.S. Government retains a nonexclusive, paid-up, irrevocable, worldwide license to publish or reproduce the published form of this manuscript or allow others to do so, for U.S. Government purposes.

## Data Availability Statement

The datasets generated and/or analyzed during the current study are available in the Material Cloud repository [39] and can be accessed at https://doi.org/10.24435/materialscloud:gh-za. In addition, the datasets are described in detail in the article by Lee et al. [26]. The underlying code for this study is publicly available at https://github.com/avakanski/Charpy_Test_ML-based_Correlation.

# References

[1] ASTM International. *ASTM E23-23: Standard test methods for notched bar impact testing of metallic materials*. (2023).

[2] Wallin, K. *Fracture toughness of engineering materials: Estimation and application*. (EMAS Publishing, 2011).

[3] Gurovich, B. A., Kuleshova, E. A., Shtrombakh, Y. I., Erak, D. Y., Chernobaeva, A. A. & Zabusov, O. O. Fine structure behaviour of VVER-1000 RPV materials under irradiation. *J. Nucl. Mater.* **389**, 490–496 (2009). https://doi.org/10.1016/j.jnucmat.2009.02.002

[4] ASTM International. *ASTM E185-19: Standard practice for design of surveillance programs for light-water moderated nuclear power reactor vessels*. (2025).

[5] Schubert, L. E., Kumar, A. S., Rosinki, S. T. & Hamilton, M. L. Effect of specimen size on the impact properties of neutron irradiated A533B steel. *J. Nucl. Mater.* **225**, 231–237 (1995). https://doi.org/10.1016/0022-3115(95)00052-6

[6] Sokolov, M. A. & Nanstad, R. K. On impact testing of subsize Charpy V-notch type specimens. In Gelles, D. S. et al. (eds) *Effects of radiation on materials: 17th International Symposium (ASTM STP 1270)* 384–409, ASTM International (1996). https://doi.org/10.1520/STP16486S

[7] Cicero, S. et al. Fracture mechanics testing of irradiated RPV steels by means of sub-sized specimens: FRACTESUS project. *Proc. Struct. Integr.* **28**, 61–66 (2020). https://doi.org/10.1016/j.prostr.2020.10.008

[8] Uehira, A. & Ukai, S. Empirical correlation of specimen size effects in Charpy impact properties of 11Cr-0.5Mo-2W, V, Nb ferritic-martensitic stainless steel. *J. Nucl. Sci. Technol.* **41**, 973–980 (2004). https://doi.org/10.1080/18811248.2004.9726320

[9] Klausnitzer, E. N. & Hofmann, G. Reconstituted Impact Specimens with Small. Inserts, *Effects of Radiation on Materials: 15th International Symposium (ASTM STP 1125)*, 76–90, ASTM International (1992). https://doi.org/10.1520/STP17863S

[10] Kumar, K. et al. Use of miniature tensile specimen for measurement of mechanical properties. *Procedia Eng.* **86**, 899–909 (2014). https://doi.org/10.1016/j.proeng.2014.11.112

[11] Bergonzi, L., Vettori, M. & Pirondi, A. Development of a miniaturized specimen to perform uniaxial tensile tests on high-performance materials. *Proc. Struct. Integr.* **24**, 213–224 (2019). https://doi.org/10.1016/j.prostr.2020.02.018

[12] Zhang, L., Harrison, W., Yar, M., Brown, S. & Lavery, N. The development of miniature tensile specimens with non-standard aspect and slimness ratios for rapid alloy prototyping processes. *J. Mater. Res. Technol.* **15**, 1830–1843 (2021). https://doi.org/10.1016/j.jmrt.2021.09.029

[13] Field, K. G., Gussev, M. N., Yamamoto, Y. & Snead, L. L. Deformation behavior of laser welds in high temperature oxidation resistant Fe–Cr–Al alloys for fuel cladding applications. *J. Nucl. Mater.* **454**, 352–358 (2014). https://doi.org/10.1016/j.jnucmat.2014.08.013

[14] Jia, W. et al. Study on intrinsic influence law of specimen size and loading speed on Charpy impact test. *Mater.* **15**, 3855 (2022). https://doi.org/10.3390/ma15113855

[15] Wallin, K., Karjalainen-Roikonen, P. & Suikkanen, P. Sub-sized CVN specimen conversion methodology (Report No. INL/EXT-08-14108). *Proc. Struct. Integr.* **2**, 3735–3742 (2016). https://doi.org/10.1016/j.prostr.2016.06.464

[16] Lucas, G. E., Odette, G. R., Sheckherd, J. W., McConnell, P. & Perrin, J. Subsized bend and Charpy V-notch specimens for irradiated testing. *ASTM Spec. Tech. Publ.* **888**, 369–387 (1986).

[17] Lucas, G. E., Odette, G. R., Sheckherd, J. W. & Krishnadev, M. R. Recent progress in subsized Charpy impact specimen testing for fusion reactor materials development. *Fusion Technol.* **10**, 728–733 (1986). https://doi.org/10.13182/FST86-A24827

[18] Corwin, W. R., Klueh, R. L. & Vitek, J. M. Effect of specimen size and nickel content on the impact properties of 12 Cr-1 MoVW ferritic steel. *J. Nucl. Mater.* **122**, 343–348 (1984). https://doi.org/10.1016/0022-3115(84)90622-6

[19] Louden, B. S., Kumar, A. S., Garner, F. A., Hamilton, M. L. & Hu, W. L. The influence of specimen size on Charpy impact testing of unirradiated HT-9. *J. Nucl. Mater.* **155–157**, 662–667 (1988). https://doi.org/10.1016/0022-3115(88)90391-1

[20] Kayano, H. et al. Charpy impact testing using miniature specimens and its application to the study of irradiation behavior of low-activation ferritic steels. *J. Nucl. Mater.* **179–181**, 425–428 (1991). https://doi.org/10.1016/0022-3115(91)90115-N

[21] Kumar, A. S., Rosinski, S. T., Cannon, N. S. & Hamilton, M. L. Subsize specimen testing of a nuclear reactor pressure vessel material. *ASTM Spec. Tech. Publ.* **1175**, 1344–1352 (1994).

[22] Wallin, K. Upper shelf energy normalisation for sub-sized Charpy V-notch specimens. *Int. J. Press. Vessels Pip.* **78**, 755–765 (2001). https://doi.org/10.1016/S0308-0161(01)00063-1

[23] Sainte Catherine, C., Hourdequin, N., Galon, P. & Forget, P. Finite element simulations of Charpy-V and sub-size Charpy tests for a low alloy RPV ferritic steel. *13th European Conf. Fracture* (ECF13) (2000). https://doi.org/10.13140/RG.2.1.3163.5045

[24] Towers, O. L. *Charpy V-notch tests: Influences of striker geometry and of specimen thickness* (Research Report 219/1983, The Welding Institute, 1983).

[25] Lucon, E., McCowan, C. N. & Santoyo, R. L. Impact characterization of line pipe steels by means of standard, sub-size and miniaturized Charpy specimens. *NIST Tech. Note* **1865**, 1–85 (2015). https://doi.org/10.6028/NIST.TN.1865

[26] Lee, I. et al. Comprehensive toughness dataset of nuclear reactor structural materials using Charpy V-notch impact testing. *Sci. Data* **12**, 543 (2025). https://doi.org/10.1038/s41597-025-04823-1

[27] Ahmedabadi, P. M. Comparison of mathematical and supervised machine-learning models for ductile-to-brittle transition in bcc alloys. *Mater. Res. Express* **11**, 076506 (2024). https://doi.org/10.1088/2053-1591/ad5cd8

[28] Chen, Y. et al. Identifying facile material descriptors for Charpy impact toughness in low-alloy steel via machine learning. *J. Mater. Sci. Technol.* **132**, 213–222 (2023). https://doi.org/2010.1016/j.jmst.2022.05.051

[29] Zhang, Y., Li, B., Li, S. et al. Using machine learning methods to predict the ductile-to-brittle transition temperature shift in RPV steel under different pulse current parameters. *Acta Metall. Sin. (Engl. Lett.)* **38**, 1029–1040 (2025). https://doi.org/10.1007/s40195-025-01840-2

[30] He, W. et al. Study on irradiation embrittlement behavior of reactor pressure vessels by machine learning methods. *Ann. Nucl. Energy* **2023**, 109965 (2023). https://doi.org/10.1016/20j.anucene.2023.109965

[31] Luo, Y. W. et al. Pore-affected fatigue life scattering and prediction of additively manufactured Inconel 718: An investigation based on miniature specimen testing and machine learning approach. *Mater. Sci. Eng. A* **802**, 140693 (2021). https://doi.org/10.1016/j.msea.2020.140693

[32] Wan, H. Y. et al. Data-driven evaluation of fatigue performance of additive manufactured parts using miniature specimens. *J. Mater. Sci. Technol.* **35**, 1137–1146 (2019). https://doi.org/2010.1016/j.jmst.2018.12.011

[33] Li, L. et al. Empirical validation of size effects in sub-sized tensile specimens for nuclear structural materials. *Sci. Rep.* **15**, 1–24 (2025). https://doi.org/10.1038/s41598-025-98849-5

[34] Pan, H. et al. Prediction of mechanical properties for typical pressure vessel steels by small punch test combined with machine learning. *Int. J. Press. Vessels Pip.* **206**, 105060 (2023). https://doi.org/10.1016/j.ijpvp.2023.105060

[35] Zhang, W. Technical problem identification for the failures of the Liberty Ships. *Challenges*, **7**(2), 20. (2016). https://doi.org/10.3390/challe7020020

[36] ASTM International. *ASTM A370-24: Standard test methods and definitions for mechanical testing of steel products*. (2024).

[37] Wallin, K. & Karjalainen-Roikonen, P. Sub-sized and miniature CVN specimen conversion methodology. *ASTM STP 1625*, 292–310 (2020). https://doi.org/10.1016/j.ijpvp.2020.104080

[38] Kurishita, H., Narui, M., Yamazaki, M., Kayano, H. & Saito, S. Effects of V-notch dimensions on Charpy impact test results for differently sized miniature specimens of ferritic steel. *Mater. Trans. JIM* **34**, 1042–1052 (1993). https://doi.org/10.2320/matertrans1989.34.1042

[39] Lee, I. et al. Dataset of Charpy V-notch impact test records of nuclear structural materials. *Mater. Cloud Arch.* **2024.206** (2024). https://doi.org/10.24435/materialscloud:gh-za

[40] Kim, B. J., Mitsui, H., Kasada, R. & Kimura, A. Evaluation of impact properties of weld joint of reactor pressure vessel steels with the use of miniaturized specimens. *J. Nucl. Sci. Technol.* **49**, 618–631 (2012). https://doi.org/10.1080/00223131.2012.687235

[41] Dohi, K., Soneda, N., Onchi, T. & Matsui, H. Correlation between subsize and full-size Charpy impact properties of neutron-irradiated reactor pressure vessel steels. In Sokolov, M. et al. (eds) *Small specimen test techniques: Fourth volume* (ASTM International, 2002). https://doi.org/10.1520/STP10818S

[42] Yuya, H. et al. Microstructure and mechanical properties of HAZ of RPVS clad with duplex stainless steel. *J. Nucl. Mater.* **545**, 152756 (2021). https://doi.org/10.1016/j.jnucmat.202020.152756

[43] Sakai, Y., Tamanoi, K. & Ogura, N. Application of tanh curve fit analysis to fracture toughness data of Japanese RPVS. *Nucl. Eng. Des.* **115**, 31–39 (1989). https://doi.org/10.1016/0029-5493(89)90257-4

[44] Lundberg, S.M. & Lee, S.-I. A unified approach to interpreting model predictions. *Advances in Neural Information Processing Systems* 30, 4765-4774 (2017).

[45] Hawthorne, J.R. *Postirradiation dynamic tear and Charpy-V performance of 12-in. thick A533-B steel plates and weld metal*. *Nucl. Eng. Des.* **17**, 116–130 (1971). https://doi.org/10.1016/0029-5493(71)90044-6

[46] Wullaert, R.A., Sheckherd, J.W. & Smith, R.W. Evaluation of the Maine Yankee reactor beltline materials. In *Irradiation effects on the microstructure and properties of metals* (ed. Shober, F.R.), ASTM International, *STP611*, 18 pp. (1976). https://doi.org/10.1520/STP38062S
[47] Server, W.L. (ed.), Wullaert, R.A., Odette, G.R. & Oldfield, W. Analysis of radiation embrittlement reference toughness curves. EPRI Report NP-1661, Research Project 886-1, Fracture Control Corporation, 340 South Kellogg Avenue, Goleta, CA (1981). https://doi.org/10.2172/6609271
[48] Hawthorne, J.R. Irradiation embrittlement. *Treatise on Mater. Sci. Technol.* **25**, 461–524 (1983). https://doi.org/10.1016/B978-0-12-341825-8.50016-7
[49] Toth, L., Rossmanith, H.-P. & Siewert, T.A. Historical background and development of the Charpy test. *Eur. Struct. Integr. Soc.* **30**, 3–19 (2002). https://doi.org/10.1016/S1566-1369(02)80002-4
[50] Bika, D., Pfaendtner, J.A., Menyhard, M. & McMahon, C.J. Jr. Sulfur-induced dynamic embrittlement in a low-alloy steel. *Acta Metall. Mater.* **43**, 1895–1908 (1995). https://doi.org/10.1016/0956-7151(94)00388-X
[51] Rudyuk, S.I., Feldman, E.I., Chernov, E.I. et al. Effect of sulfur and phosphorus on the properties of steel 18B. *Met. Sci. Heat Treat.* **16**, 1056–1059 (1974). https://doi.org/10.1007/BF00664052
[52] Wang, X., Du, Y., Zhang, R. et al. Effects of nickel contents on phase constituent and mechanical properties of ductile iron. *Int. J. Met. Cast.* **18**, 2721–2731 (2024). https://doi.org/10.1007/s40962-023-01208-1
[53] Splett, J.D., Iyer, H.K., Wang, C.-M. & McCowan, C.N. *NIST recommended practice guide: Computing uncertainty for Charpy impact machine test results*. NIST Special Publication 960-18, National Institute of Standards and Technology, Boulder, CO (2007). http://dx.doi.org/10.13140/RG.2.1.4476.9683
[54] Lont, M.A. The determination of uncertainties in Charpy impact testing. In Manual of codes of practice for the determination of uncertainties in mechanical tests on metallic materials, Code of Practice No. 06, Standards Measurement & Testing Project No. SMT4-CT97-2165, TNO Institute of Industrial Technology, Apeldoorn, The Netherlands (2000).
[55] Splett, J.D., Wang, C.-M. & McCowan, C.N. *Charpy machine verification: Limits and uncertainty*. NIST Special Publication 260-171, National Institute of Standards and Technology, Boulder, CO (2008).
[56] Todinov, M. T. Uncertainty and risk associated with the Charpy impact energy of multi-run welds. *Nucl. Eng. Des.* **231**, 27–38 (2004). https://doi.org/10.1016/j.nucengdes.2004.02.003
[57] Awachat, P. & Dakre, V. *Analysis of various parameters responsible for measurement uncertainty in Charpy impact testing*. *AIP Conf. Proc.* **2417**, 050002 (2021). https://doi.org/10.1063/5.0072681
[58] Marini, B. Empirical estimation of uncertainties of Charpy impact testing transition temperatures for an RPV steel. *EPJ Nucl. Sci. Technol.* **6**, 57 (2020). https://doi.org/10.1051/epjn/2020019
[59] Takamizawa, H., Nishiyama, Y. & Hirano, T. Bayesian uncertainty evaluation of Charpy ductile-to-brittle transition temperature for reactor pressure vessel steels. In *ASME 2020 Pressure Vessels & Piping Conference*, Vol. 1: Codes and Standards, PVP2020-21698 (2020). https://doi.org/10.1115/PVP2020-21698
[60] Stallmann, F. *Curve fitting and uncertainty analysis of Charpy impact data*. Oak Ridge National Laboratory (1982).
[61] Li, L., Chang, J., Vakanski, A., Wang, Y., Yao, T. & Xian, M. Uncertainty quantification in multivariable regression for material property prediction with Bayesian neural networks. *Sci. Rep.* **14**, 10543 (2024). https://doi.org/10.1038/s41598-024-61189-x